*Accepted version Dec 2024 – To Appear in Expert Systems with Applications*

# Using GPT-4 to guide causal machine learning

Anthony C. Constantinou[1], Neville K. Kitson[1], and Alessio Zanga[1,2,3]

1. Bayesian AI research lab, Machine Intelligence and Decision Systems (MInDS) research group, School of Electronic Engineering and Computer Science, Queen Mary University of London, E1 4NS, London, United Kingdom.
2. Models and Algorithms for Data and Text Mining Laboratory (MADLab), Department of Informatics, Systems and Communication, University of Milano - Bicocca, Milan, Italy
3. Data Science and Advanced Analytics, F. Hoffmann - La Roche Ltd, Basel, Switzerland

E-mails: a.constantinou@qmul.ac.uk, n.k.kitson@qmul.ac.uk, and alessio.zanga@unimib.it.

**Abstract:** Since its introduction to the public, ChatGPT has had an unprecedented impact. While some experts praised AI advancements and highlighted their potential risks, others have been critical about the accuracy and usefulness of Large Language Models (LLMs). In this paper, we are interested in the ability of LLMs to identify causal relationships. We focus on the well-established GPT-4 (Turbo) and evaluate its performance under the most restrictive conditions, by isolating its ability to infer causal relationships based solely on the variable labels without being given any other context by humans, demonstrating the minimum level of effectiveness one can expect when it is provided with label-only information. We show that questionnaire participants judge the GPT-4 graphs as the most accurate in the evaluated categories, closely followed by knowledge graphs constructed by domain experts, with causal Machine Learning (ML) far behind. We use these results to highlight the important limitation of causal ML, which often produces causal graphs that violate common sense, affecting trust in them. However, we show that pairing GPT-4 with causal ML overcomes this limitation, resulting in graphical structures learnt from real data that align more closely with those identified by domain experts, compared to structures learnt by causal ML alone. Overall, our findings suggest that despite GPT-4 *not* being explicitly designed to reason causally, it can still be a valuable tool for causal representation, as it improves the causal discovery process of causal ML algorithms that *are* designed to do just that.

## 1. Introduction

Causal discovery moves beyond mere correlations to uncover the underlying causal mechanisms that drive observed phenomena. Determining a causal graph enables the parameterisation of causal models, such as a Causal Bayesian Network (CBN), which can then be used for causal inference and optimal decision-making under uncertainty through simulation of hypothetical interventions. A CBN is a probabilistic graphical model represented by a Directed Acyclic Graph (DAG), where nodes represent variables, and directed edges indicate causal relationships between these variables. Each node in a CBN is described by a Conditional Probability Distribution (CPD) that quantifies the effect of its parent nodes. This structure allows for the representation of complex causal relationships and the computation of joint conditional and marginal probability distributions.

A CBN supports both backward and forward inference. For example, predicting effects such as symptoms given a cause such as disease, or inferring the most likely disease cause given observed symptoms. More importantly, causal models enable the simulation of hypothetical interventions and estimation of their effects before real-world implementation, which is crucial for decision support. For a comprehensive review of causal Machine Learning (ML) algorithms, we direct readers to Kitson et al. (2023) and Zanga et al. (2022).

Despite their utility, causal ML algorithms face significant challenges that necessitate combining these algorithms with domain knowledge or interventional data. Three key limitations that are relevant to this study are:

a. **Uncertainty in the number of edges:** Causal ML algorithms often face significant challenges in accurately recovering the correct number of edges in a causal graph. One major limitation is their tendency to underestimate the number of edges when the sample size is low. This occurs because limited data can obscure subtle dependencies and causal relationships, leading to an unreasonably sparse graph. Conversely, when the sample size is high, these algorithms may overestimate the number of edges, often due to overfitting issues or inability to disentangle all spurious relationships from causal relationships. Consequently, the reliability of these algorithms can vary significantly with the sample size, impacting their effectiveness in accurately modelling causal structures.

b. **Incomplete orientation of edges:** Causal ML algorithms typically do not orientate all the edges they discover. This limitation arises because observational data alone is generally insufficient to distinguish between different causal graphs, often requiring either interventional (also refer to as experimental) data for complete causal discovery, or additional strong assumptions which force edge orientations from observational data. In the absence of additional assumptions that force edge orientations irrespective of the input data, a causal ML algorithm typically employs an objective function that is score-equivalent, allocating the same score to any two DAG structures that are part of the same Markov Equivalence Class (MEC). A MEC of DAGs is a set of DAGs that entail the same conditional independencies, and each MEC is represented by a Completed

Partially DAG (CPDAG). A CPDAG contains both directed and undirected edges, where a directed edge indicates that all of the DAGs within the MEC have the same orientation for that specific edge, whereas an undirected edge indicates a directional inconsistency between those DAGs.

c. **Irrational orientation of edges:** Even when edges are orientated, some may be wrongly-orientated, and may even appear completely irrational to a human in that they disobey the fundamental tenets of causality. For instance, an algorithm might incorrectly suggest that *Dance moves* cause *Music*, or that *Rainbow* causes *Rain*. This is partly due to causal ML algorithms not being provided with key temporal information about the input variables; i.e., data indicating that event $A$ occurs after observing $B$, and hence the constraint that $B$ cannot cause $A$. While it has been argued that objective temporal information should form part of observational data in causal discovery (Constantinou, 2021), it is generally viewed as a form of optional subjective information that is overlooked, contributing to these orientational inaccuracies.

Large Language Models (LLMs) represent a class of artificial intelligence models designed to understand and generate human-like text. These models are built on deep learning architectures, particularly transformers, which enable them to process and produce natural language text with remarkable accuracy and fluency. The most well-known example is OpenAI's ChatGPT, with iterations like GPT-2, GPT-3, and beyond setting new benchmarks for generating coherent and contextually relevant text. LLMs are not designed to disentangle correlation from causation, and some argue that it is crucial for LLMs to reason causally in order to generate logical inferences. Because LLMs do not reason causally by design, significant debate remains as to whether they merely generate restructured memorised information or go beyond that and towards some form of causal reasoning (Bubeck et al., 2023; Zhong et al., 2023; Zhou et al., 2024).

Since the public release of GPT-3, there has been a growing interest in utilising LLMs for causal discovery, with studies highlighting conflicting conclusions about their causal reasoning capabilities. We begin with the papers that conclude that LLMs are mostly inadequate in terms of causal reasoning. These include Jin et al. (2024) who evaluated 17 LLMs on causal inference skills and found that these models "*achieve almost close to random performance*". Zhou et al. (2024) explored criteria for benchmarking the causal learning capabilities of LLMs and concluded that "*even the most advanced LLMs do not yet match the performance of classic and SOTA methods in causal learning*". They illustrated that while LLMs can compete with state-of-the-art (SOTA) methods when the problem relies on small datasets, their effectiveness diminishes with larger datasets. Long et al. (2024) showed that the accuracy of GPT-3 in causal discovery depends on the language used by the user to describe the relationship between two events, concluding that "*the use of LLMs to build DAGs should be, at present, only conducted with expert verification*". Zhang et al. (2023) suggested that while LLMs can answer causal questions based on existing knowledge, they are still incapable of providing satisfactory answers to problems involving new knowledge. Pawlowski et al. (2024) demonstrated that neither context-augmented LLMs, that are given the non-parameterised causal graph, nor API-augmented LLMs that are given the parameterised causal model, can correctly solve causal question-answering tasks. Tu et al. (2023)

tested the ability of ChatGPT to answer causal discovery questions about a neuropathic pain diagnosis case study, and showed that while ChatGPT is good at correctly discovering true positives, it is poor at correctly identifying false negative causal relationships. Lastly, Zečević et al. (2023) focused on experiments with Structural Causal Models (SCMs) to illustrate and argue that LLMs not only cannot reason causally, but are also weak 'causal parrots'.

In contrast to the above studies that highlight the inability of LLMs to reason causally, other research indicates that LLMs are adequate in producing causal graphs. For instance, Kiciman et al. (2023) studied the capabilities of LLMs on various causal reasoning tasks and found that algorithms based on GPT-3.5 and GPT-4 "*outperform state-of-the-art causal algorithms in graph discovery and counterfactual inference*". Lyu et al. (2022) explored the capability of LLMs in establishing causality between two variables at a time, which is not generally feasible for causal discovery algorithms that do not make additional assumptions to force orientations (i.e., some algorithms claim to be able to orientate all edges, but this requires strong assumptions about the nature of noise in the data), and showed that LLMs are effective in distinguishing cause from effect. In a similar study, Jiralerspong et al. (2024) propose a framework that, instead of performing pairwise queries to LLMs which require a quadratic number of queries with the number of variables, it conducts a breadth-first search with only a linear number of queries, demonstrating positive results on real-world graphs. Long et al. (2023) demonstrated that LLMs can serve as imperfect domain experts, helping causal discovery algorithms to select the correct DAG from a MEC, whereas Takayama et al. (2024) acknowledge the challenges associated with acquiring domain knowledge and propose an approach for eliciting causal edges from LLMs, demonstrating that GPT-4 improves data-driven causal discovery by recovering graphs that are closer to the ground truth. In a similar study, Cohrs et al. (2024) explore using LLMs as an alternative to domain experts for causal graph generation, and frame conditional independence queries as prompts to LLMs, showing that when the LLM-generated results were provided to the PC algorithm, the resulting graph was a plausible causal representation. Zhang et al. (2024) showed that pairing LLMs with Retrieval Augmented-Generation (RAG) solutions enables them to recover causal graphs that are more accurate than those learnt by causal ML, while Antonucci et al. (2023) found that LLMs are competitive in inferring causal relationships with traditional natural language processing and deep learning techniques. Lastly, Le et al. (2024) present a framework that uses the multi-agent capabilities of LLMs for causal reasoning, and demonstrate how causal-related problems could be addressed by combining reasoning skills with statistical analysis through multi-agent LLM collaboration. For a detailed review on integrating LLMs with causal discovery, we refer readers to a recent survey by Wan et al. (2024).

In this paper, we use a questionnaire to gather data from human participants on their ability to identify whether LLMs, causal ML, or domain experts constructed the presented causal graphs, and to evaluate and comment on their accuracy. Additionally, we investigate whether the causal relationships extracted from GPT-4 (Turbo) can address some of the current limitations in causal ML, with a focus on the two limitations discussed above in incomplete and irrational edge orientations. The remainder of the paper is structured as follows: Section 2 describes the methodology and experimental setup, Section 3 presents and discusses the results, and Section 4 provides our concluding remarks, highlighting limitations and future research directions.

## 2. Methodology and experimental setup

Figure 1 illustrates the complete methodology, with descriptions provided in the subsections that follow. We have made all files needed to reproduce the results of this study, including the real datasets, graphs constructed by domain experts, GPT-4 prompts, GPT-4 outputs, GPT-4 averaged outputs, GPT-4 constraints, as well as the questionnaire responses, publicly available through the Bayesys repository (Constantinou et al., 2020).

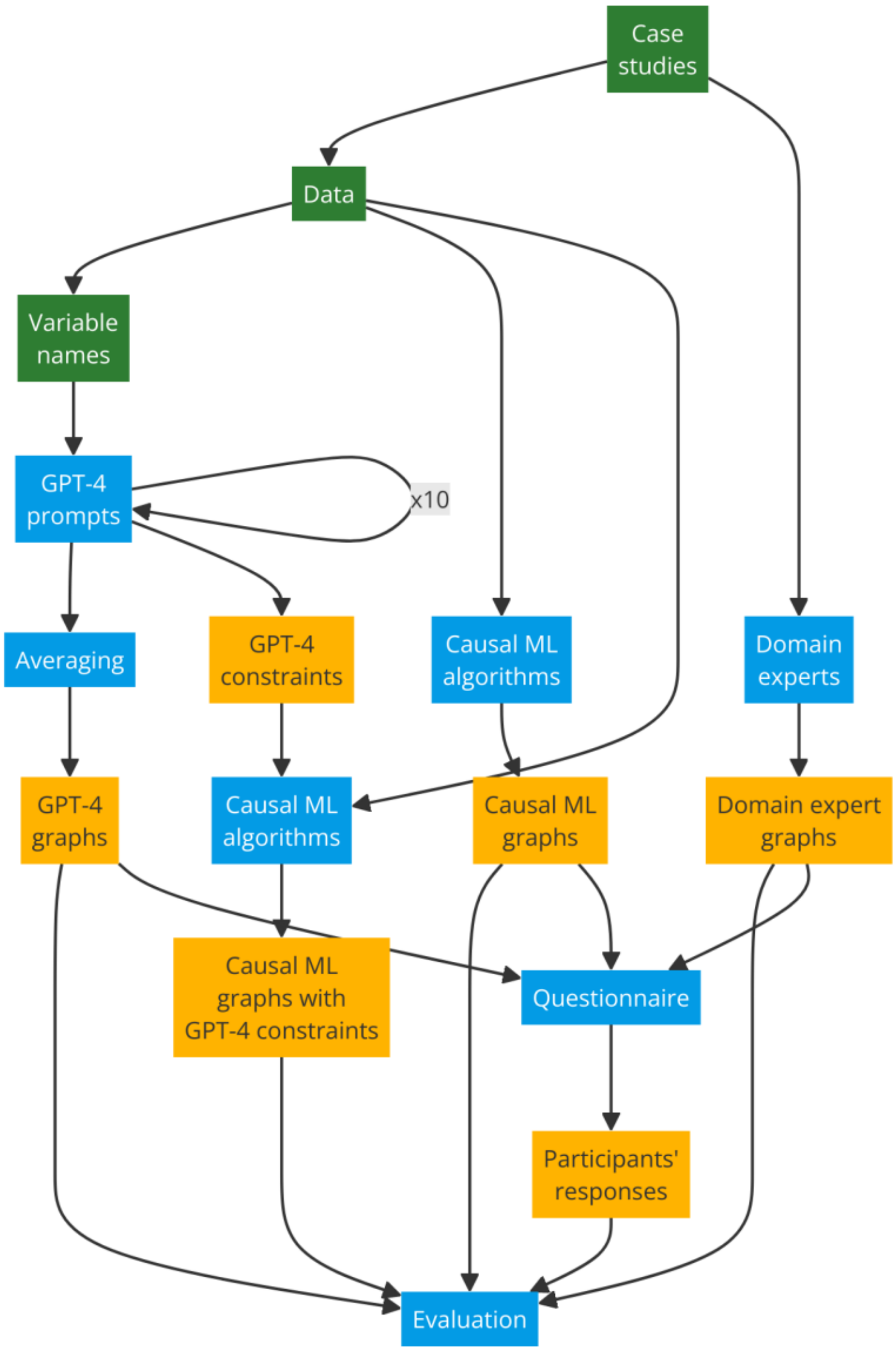

**Figure 1.** The process we followed to compare LLM graphs with causal ML and domain expert graphs, and evaluate the participants' responses, where green nodes can be viewed as inputs, blue nodes as processes, and orange nodes as outputs.

### *2.1. Case studies as input to GPT-4 (Turbo)*

Five case studies were selected from diverse domains for a more comprehensive evaluation. These are detailed in Table 1, which shows that the case studies vary widely across several dimensions: domain, variable size (ranging from 9 to 56), sample size (from under a thousand to hundreds of thousands), number of edges (from 15 to 95), and free parameters (from approximately a thousand to around 39 million). The networks also differ in graph complexity, with maximum in-degree (number of parents) ranging from 2 to 17, maximum out-degree (number of children) from 6 to 15, and maximum degree (number of neighbouring nodes) from 6 to 22.

We avoided selecting case studies incorporating hundreds of variables to ensure that a) the case studies are simple enough to enable questionnaire participants to review them, and b) the number of the variable labels can be processed by GPT-4, since there is a limit to the number of characters an input to LLMs can have, which varies with platforms and implementation versions. The selected case studies span various domains, outlined as follows:

a. Sports: A small BN model that combines football team ratings with various performance statistics to predict different match outcomes.
b. COVID-19: A medium-sized BN model capturing key events of the COVID-19 pandemic in the UK, including mobility measures and vaccination efforts, and their influence on infection rates and hospitalisations.
c. Property: A medium-sized BN model developed to assess investment decisions within the UK property market. Because this case study lacks real data, it is included in the questionnaire analyses but not in the causal ML evaluations.
d. Diarrhoea: A medium-sized BN model investigating factors associated with childhood diarrhoea in India, using data from a large demographic and health survey.
e. ForMed: A large BN model that evaluates the risk of violent reoffending amongst mentally ill prisoners and examines potential interventions available to probation officers for managing this risk.

**Table 1.** The case studies used to evaluate the causal reasoning of GPT-4. All five case studies are taken from the Bayesys repository (Constantinou et al., 2020).

| Case study | Complexity of the real dataset | | | Complexity of the knowledge DAG | | | | |
|---|---|---|---|---|---|---|---|---|
| | Variables | Sample size | Model/Data type | Edges | Free parameters | Max in-degree | Max out-degree | Max degree |
| Sports | 9 | 3,536 | Discrete | 15 | 1,049 | 2 | 7 | 7 |
| COVID-19 | 17 | 866 | Discrete | 37 | 7,834 | 5 | 7 | 10 |
| Property | 27 | n/a | Discrete | 31 | 3,056 | 3 | 6 | 6 |
| Diarrhoea | 28 | 259,627 | Discrete | 68 | 1716 | 8 | 15 | 17 |
| ForMed | 56 | 953 | Discrete | 95 | 39,196,846 | 17 | 11 | 22 |

a. **Input preparation for GPT-4:** For each case study, we provide the labels of the variables as input to GPT-4, and ask GPT-4 to identify causal relationships between the labels. Specifically, GPT-4 was asked to specify a set of directed edges representing causal links between the input variables. No additional context or data was given to GPT-4, isolating its ability to infer causal relationships based solely on the labels. Moreover, because the way a question is posed to GPT-4 may influence its output (noting that even identical questions often generate slightly different responses), we repeated this process 10 times for each case study using different prompts generated by GPT-4, as shown in Table 2.

**Table 2.** The 10 prompts we used to ask GPT-4 to generate causal relationships between the input variables for each case study. These prompts were obtained from GPT-4 using the following prompt "*Generate 10 different ways to ask someone to provide a list of causal relationships between variables in a given dataset*".

| Prompt no. | Prompt |
|---|---|
| 1 | "*Could you identify and list the causal connections among the variables within the dataset?*" |
| 2 | "*Would you mind detailing the cause-and-effect relationships present among the dataset's variables?*" |
| 3 | "*Can you provide an analysis of the causal linkages between the dataset's variables?*" |
| 4 | "*I'd appreciate it if you could enumerate the causative associations among the variables in our dataset.*" |
| 5 | "*Could you explore and list out the causal relations found within the dataset's variables?*" |
| 6 | "*Please, could you dissect and document the causal connections that exist among the dataset's variables?*" |
| 7 | "*Would you be able to chart out the causal pathways linking the variables in the dataset?*" |
| 8 | "*Can you draft a list of causal relationships that are evident among the variables of the dataset?*" |
| 9 | "*I'd like you to investigate and compile a list of the cause-and-effect dynamics among the dataset's variables.*" |
| 10 | "*Could you analyze and itemize the causal links present within the dataset, focusing on the variables' interactions?*" |

### *2.2. Questionnaire*

A questionnaire was designed for human participants to evaluate the different causal graphs produced by GPT-4 based on variable labels, causal ML based on data samples, and domain experts based on their subjective causal knowledge. A sample of the questionnaire is shown in Figure A.1, showing the first causal graph of the first case study. Participants were free to complete one or up to all five case studies. It was completely up to them to decide how many, and which, of the case studies they completed. This option was necessary to ensure that we did not force participants to complete case studies they were not be able to judge reasonably well. Moreover, we estimated that each case study required an average of 6 minutes to complete, which makes for a total of 30 minutes for those who decide to complete the questionnaire in full.

The questionnaire involved three causal graphs for each case study in Table 1, for a total of 15 causal graphs. The three graphs for each case study represent the following:

a. ***Knowledge graphs*:** These are the causal graphs elicited from domain experts. They are taken from the Bayesys repository (Constantinou, 2020), and are based on the knowledge graphs as published in the original studies.

b. ***Causal ML graphs*:** These are causal graphs learnt with causal ML algorithms from real case study data. For the *Diarrhoea* and *COVID-19* case studies, we took the learnt graphs from the original studies. The other three studies did not employ causal ML, so these graphs were not available. We, therefore, learnt the structures using and a set of algorithms spanning different classes of learning; i.e., score-based HC, Tabu, GES and MAHC, constraint-based PC-Stable, and hybrid MMHC and SaiyanH.

   However, because our aim here was to obtain a single DAG structure representative of causal ML, we performed model-averaging on the set of causal ML graphs learnt for each case study. We use a model-averaging process similar to (Petrungaro et al., 2024; Zahoor et al., 2024; Constantinou et al., 2023a), where the average graph contains all the edges that appear in at least two thirds of the graphs in the input set of learnt graphs, as long as an edge added to the average graph - starting from the directed edges that appear the most times within the set of graphs - does not produce a cycle.

c. ***LLM graphs*****:** These are the causal graphs obtained by GPT-4 as described in subsection 2.1. Because we obtained 10 graphs per case study, we applied the same model-averaging process described in (b) above in order to retrieve a single DAG structure for each case study that is representative of the GPT-4 output.

Participants were shown the causal graphs and asked to specify whether they had been produced from domain knowledge, causal ML, or LLM. They were also asked to judge the accuracy of each graph. Answering these questions involved selecting one of four possible responses:

a. *Very Likely*, *Likely*, *Unlikely*, and *Very Unlikely*, in determining whether a graph was constructed by human experts, causal ML, or LLM;
b. *Very Accurate*, *Mostly Accurate*, *Mostly Inaccurate*, and *Very Inaccurate*, in judging the accuracy of a causal graph.

The participants were also given the option to comment on each graph presented to them. Key comments left by participants are presented in Table 7 and are discussed in Section 3.

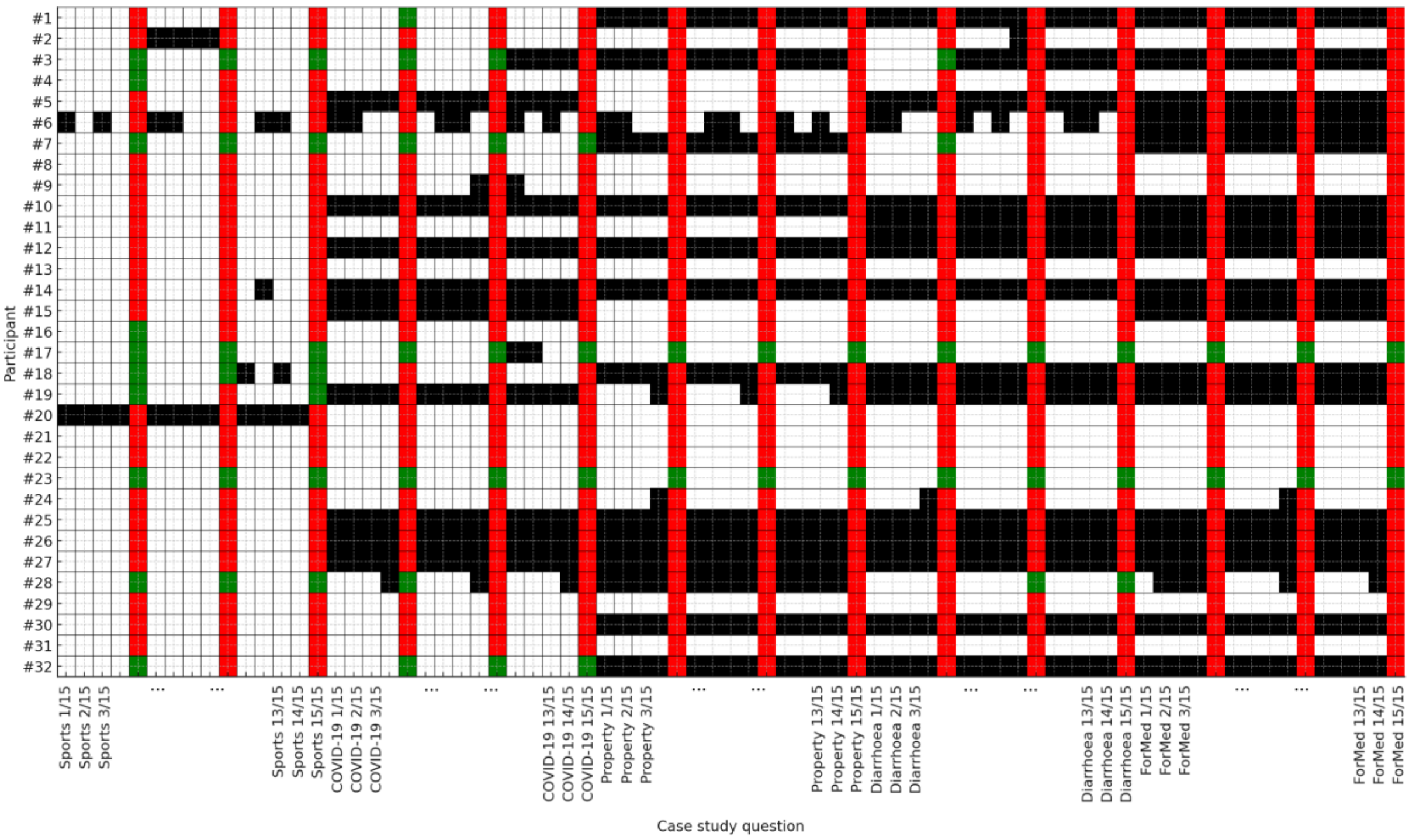

**Figure 2.** Visualisation of participants' responses, where white and black boxes indicate a response or no-response to multiple-choice questions, while green and red boxes indicate a response or no-response to optional free-text comment questions.

Figure 2 presents a box visualisation showing participants' responses to multiple-choice and free-text comment questions across each case study. The results indicate higher participant engagement in the earlier, smaller, case studies compared to the later ones which are larger networks, possibly due to increasing complexity in each subsequent case study. Specifically, participants completed 93%, 67%, 57%, 55%, and 45% of the 15 multiple-choice questions per case study, for the Sports, COVID-19, Property, Diarrhoea, and ForMed case studies, respectively. For the optional free-text comment questions, response rates were 24%, 17%, 6%, 10%, and 6% for each case study in the same order.

### *2.3. Using LLMs to guide causal ML*

Unlike the common practice of evaluating causal ML algorithms with synthetic data due to the absence of real-world ground truth graphs, this paper investigates whether LLMs can assist causal ML in learning graphs that more closely align with those constructed by domain experts. Specifically, we investigate the usefulness of the causal relationships generated by GPT-4 in terms of guiding causal ML algorithms when learning from real data. We employ a systematic approach that involves multiple algorithms across different classes of structure learning, and test those algorithms on real case-study data with and without GPT-4 constraints, with different quantities of constraints.

We begin by describing how we convert GPT-4 outputs into constraints. As discussed in Section 2.1, the variable labels are provided as input to GPT-4 using 10 different prompts, leading to 10 GPT-4 outputs. We take those 10 outputs, for each case study, and record the edges that appear in at least a third (33%), a half (50%), and two thirds (67%) of those 10 outputs. These differing number of edges are reflected by the different numbers of constraints, so that we assess the robustness and consistency of the causal relationships proposed by GPT-4 across different levels of confidence. Table 3 presents the results by repeating this across all five case studies, leading to 15 different sets of edges. It is worth noting that for the ForMed case study, no same edge appeared in at least two-thirds of the 10 outputs, resulting in an empty generated edge set for that experiment. This may be due to the relatively large number of nodes in the ForMed network (56), suggesting – together with the results in Table 3 - that the output of GPT-4 becomes increasingly distorted with variable size, leading to greater inconsistency between outputs.

**Table 3.** The number of edge-sets extracted from GPT-4 for each case study, based on the specified threshold rates about the proportion of times the same edge appeared across each of the 10 GPT-4 prompts per case study.

| Case study | Edges (rate 33%) | Edges (rate 50%) | Edges (rate 67%) |
|---|---|---|---|
| Sports | 14 | 14 | 8 |
| Covid-19 | 27 | 20 | 13 |
| Diarrhoea | 34 | 25 | 9 |
| ForMed | 32 | 7 | 0 |

We then take each set of edges specified in Table 3, and convert it into three different types of constraints that could be used to guide structure learning algorithms. These three types of constraints, which are described in (Constantinou et al., 2023b) in greater technical detail, are:

a. **Required edges**: explicitly define the directionality of causal links between variables, where the search space of graphs is restricted to structures containing the specified directed edges.

b. **Initial graph:** also known as starting graph, represents the starting point for exploration in the search-space of graphs. For most algorithms, the starting point is typically an empty, a fully connected, or a random structure. When the set of constraints is given as an initial graph, then the starting point in the search space is the structure specified in the set of constraints.

c. **Temporal order:** also known as temporal edge tiers, ensures that the temporal order of events was respected, preventing causal directions that contradict temporal sequences. Specifically, the search space of graphs explored is restricted to graphical structures that satisfy the temporal constraints, converted from the set of required edge constraints. For example, if $A \rightarrow B$ and $B \rightarrow C$ appear in a set of constraints, these two edges alone would produce multiple temporal constraints; i.e., $B$ cannot a parent nor an ancestor of $A$ (although not all implementations impose restrictions on ancestral relationships), and $C$ cannot be a parent nor an ancestor of neither $A$ nor $B$. In this case, the search-space of graphical structures is restricted to DAGs that do not violate any of the temporal orderings implied by the set of required edge constraints.

To enforce these constraints, we selected algorithmic implementations that support structure learning with constraints on discrete data. Table 4 lists these algorithms, their class of learning and implementation details. We used the constraint-based PC-Stable algorithm, the score-based Fast Greedy Equivalence Search (FGES), Hill-Climbing (HC), TABU, and Model-Averaging Hill-Climbing (MAHC) algorithms, and the hybrid Max-Min Hill Climbing (MMHC) and SaiyanH algorithms.

**Table 4.** The causal ML implementations tested (Scutari, 2010 for bnlearn; Ramsey et al., 2018 for Tetrad; Constantinou, 2019 for Bayesys), that support discrete data and simulation of the specified structural constraints.

| Algorithm | Learning class | Library/ Software | Required edge constraints | Initial graph constraints | Temporal constraints |
|---|---|---|---|---|---|
| FGES | Score-based | Tetrad | Yes | No | Yes |
| HC | Score-based | Bayesys | Yes | Yes | Yes |
| MAHC | Score-based | Bayesys | Yes | Yes | Yes |
| MMHC | Hybrid | bnlearn | Yes | No | Yes |
| PC-Stable | Constraint-based | bnlearn | Yes | No | Yes |
| SaiyanH | Hybrid | Bayesys | Yes | Yes | Yes |
| TABU | Score-based | Bayesys | Yes | Yes | Yes |

## 3. Results and Discussion

The results are separated into two subsections. The first part focuses on the questionnaire outcomes, while the second part focuses on structure learning outcomes.

### *3.1. Questionnaire outcomes*

We invited approximately 200 MSc students, 300 PhD students, and 1,000 LinkedIn connections to complete the questionnaire. The 200 MSc students invited were enrolled in the post-graduate Data Analytics course at Queen Mary University of London (QMUL), where 40% of the material is based on causal ML. The 300 PhD students invited were enrolled in the School of Electronic Engineering and Computer Science at QMUL. The 1,000 LinkedIn connections invited included academics and industry professionals across different disciplines.

We received 32 responses from 11[1] different universities or organisations, resulting in a response rate of approximately 2.13%. The rather low response rate may be partly explained by the fact that this questionnaire was unfunded and so the respondents were not offered any payment for their participation. Additionally, the questionnaire took a relatively long time to complete, with an estimated six minutes per case study, resulting in a maximum total of 30 minutes for those who chose to complete all five case studies. Figure 3 shows the distribution of the responses by the participants' expertise or knowledge in a pre-determined set of domains relevant to the case studies.

Despite the rather limited number of responses, consistent patterns emerged across all five case studies. As presented in Table 5 and detailed in Figure 4, the questionnaire responses suggest that GPT-4 is the most reliable method for achieving higher accuracy in the evaluated categories, closely followed by knowledge graphs, with causal ML far behind. Specifically, as shown in Table 5, GPT-4 was consistently judged by participants as the highest for accuracy scores, with knowledge graphs generally close to those of GPT-4. The graphs learnt by causal ML, however, received the lowest accuracy across all categories.

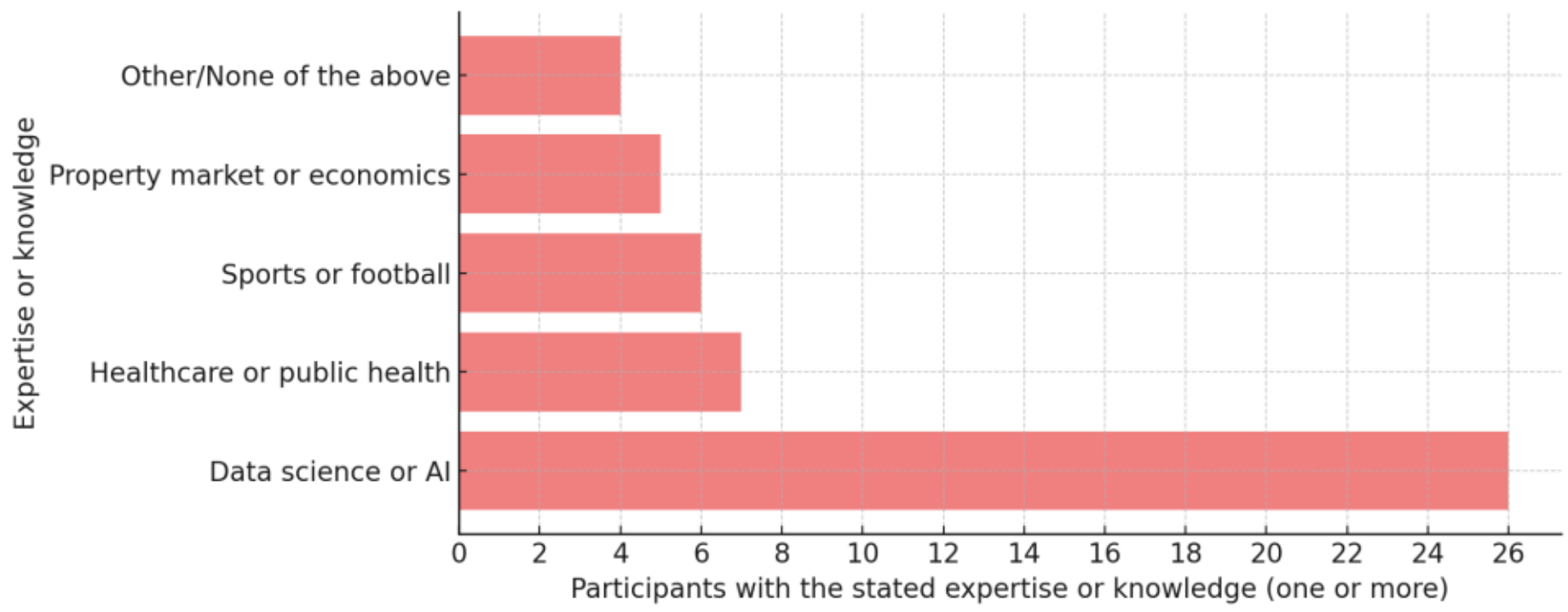

**Figure 3.** Questionnaire responses distributed by the participants' stated (one or more) expertise or skill.

These results do not necessarily suggest that causal ML is less effective than the other two methods in this context. Instead, they highlight and support the important limitation of causal ML discussed in the introduction, in that they often produce causal relationships that are counterintuitive, which is not something we would expect from a domain expert or LLM. As shown in Table 7, which presents some of the key optional comments provided by participants, most of them commented negatively on the graphs learnt by causal ML, and say that the graphical structures tend to be sparse or too simplistic, some relationships seem counterintuitive or wrong, some edge orientations appear to be incorrect, and some key relationships are missed.

[1] From Queen Mary University of London, University of Milano – Bicocca, University of Oxford, University of Toronto, Munster Technological University, Ministry of Health, Middle East Technical University, Stock exchange, Indian Institute of Science Education and Research - Bhopal, UNSW Sydney, and University of Utah.

Presumably, these observations helped the questionnaire participants to more accurately identify the graphical structures generated by causal ML. As shown in Table 6, all five causal ML graphs were correctly identified by the average participant. On the other hand, most of the knowledge graphs were incorrectly identified as LLM graphs, whereas most LLM graphs were incorrectly identified as knowledge graphs. Overall, the results suggest that the participants were accurate in identifying graphical structures learnt with causal ML, but they were partly correct in identifying knowledge graphs and LLM graphs, often confusing a knowledge graph as an LLM graph and vice-versa.

**Table 5.** The accuracy of the 15 causal graphs as determined by questionnaire responses, where $\text{Overall score} = \text{Very accurate} \times 1 + \text{Mostly accurate} \times 0.66 + \text{Mostly inaccurate} \times 0.33 + \text{Very Inaccurate} \times 0.00$.

| Graph | Very accurate | Mostly accurate | Mostly inaccurate | Very inaccurate | Overall score |
|---|---|---|---|---|---|
| Sports (Knowledge) | 26.7% | 56.7% | 16.7% | 0.0% | 69.63 |
| COVID-19 (Knowledge) | 19.0% | 57.1% | 23.8% | 0.0% | 64.54 |
| Property (Knowledge) | 0.0% | 76.5% | 23.5% | 0.0% | 58.25 |
| Diarrhoea (Knowledge) | 5.9% | 70.6% | 23.5% | 0.0% | 60.25 |
| ForMed (Knowledge) | 0.0% | 64.3% | 28.6% | 7.1% | 51.88 |
| Sports (Causal ML) | 9.7% | 29.0% | 41.9% | 19.4% | 42.67 |
| COVID-19 (Causal ML) | 0.0% | 27.3% | 68.2% | 4.5% | 40.52 |
| Property (Causal ML) | 0.0% | 44.4% | 50.0% | 5.6% | 45.80 |
| Diarrhoea (Causal ML) | 5.5% | 66.7% | 16.7% | 11.1% | 55.03 |
| ForMed (Causal ML) | 7.1% | 50.0% | 35.7% | 7.1% | 51.88 |
| Sports (LLM) | 16.1% | 80.6% | 3.3% | 0.0% | 70.39 |
| COVID-19 (LLM) | 14.3% | 76.2% | 9.5% | 0.0% | 67.73 |
| Property (LLM) | 5.5% | 77.8% | 16.7% | 0.0% | 62.36 |
| Diarrhoea (LLM) | 16.7% | 61.1% | 22.2% | 0.0% | 64.35 |
| ForMed (LLM) | 7.7% | 61.5% | 30.8% | 0.0% | 58.45 |

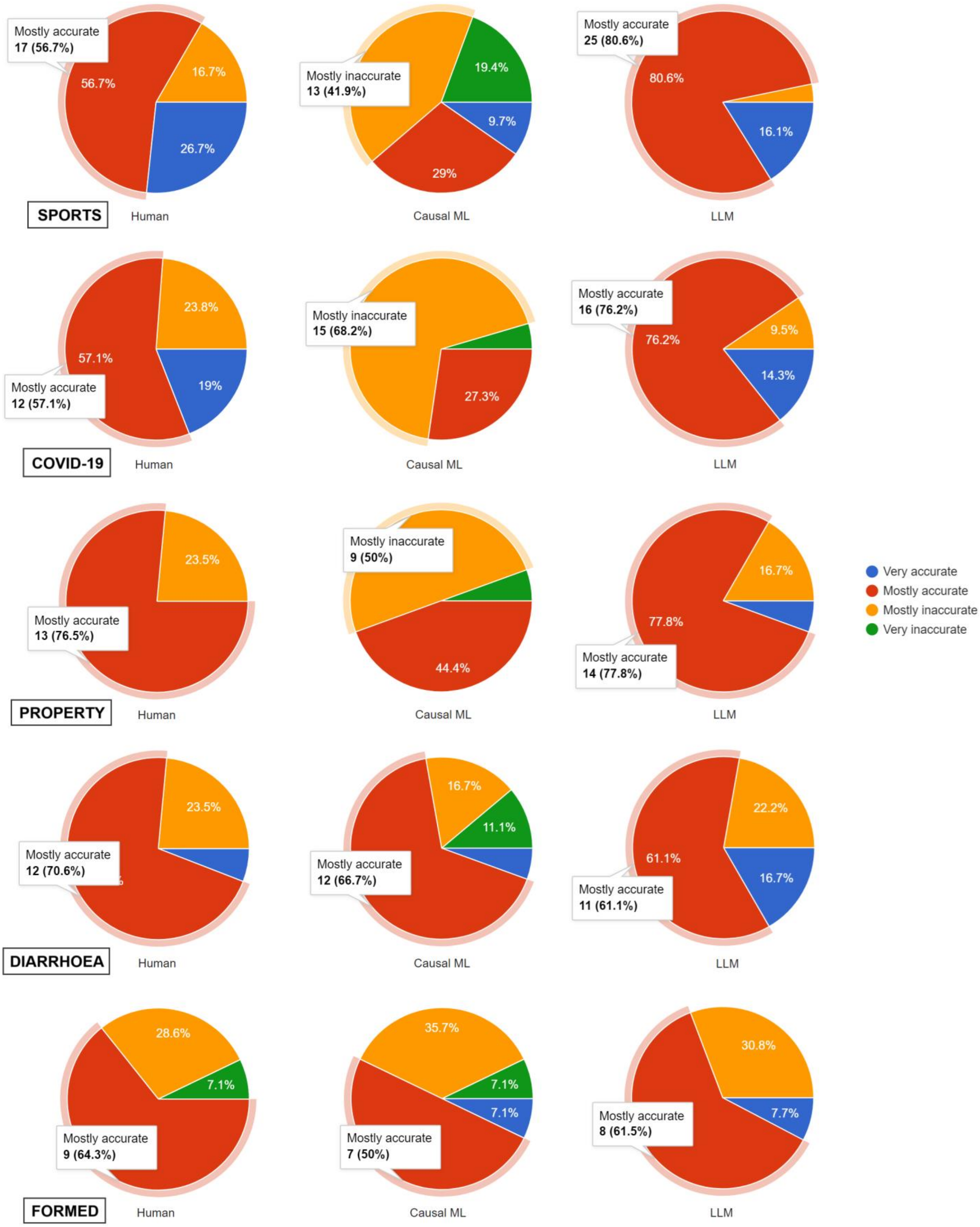

**Figure 4.** How the questionnaire participants assessed each of the 15 causal graphs in terms of causal representation accuracy.

**Table 6.** How the questionnaire participants classified each of the 15 graphs. The classification is determined by the responses presented in Figures B1, B2, and B3, where the scores presented are derived in the same way as in Table 5; i.e., $\text{Highly likely} \times 1 + \text{Likely} \times 0.66 + \text{Unlikely} \times 0.33 + \text{Highly unlikely} \times 0$.

| | Classification by participants | | | |
|---|---|---|---|---|
| **Graph** | **Knowledge** | **Causal ML** | **LLM** | **Overall** |
| Sports (Knowledge) | 19.2 | 18.3 | 19.6 | LLM |
| COVID-19 (Knowledge) | 12.3 | 15.6 | 13.3 | Causal ML |
| Property (Knowledge) | 11.3 | 10.3 | 11.9 | LLM |
| Diarrhoea (Knowledge) | 9.6 | 11.3 | 11.9 | LLM |
| ForMed (Knowledge) | 7.3 | 7.9 | 9.9 | LLM |
| Sports (Causal ML) | 11.6 | 17.9 | 17.6 | Causal ML |
| COVID-19 (Causal ML) | 8.3 | 13.6 | 12.9 | Causal ML |
| Property (Causal ML) | 7.6 | 10.9 | 10.0 | Causal ML |
| Diarrhoea (Causal ML) | 9.3 | 11.6 | 10.3 | Causal ML |
| ForMed (Causal ML) | 7.3 | 7.9 | 7.3 | Causal ML |
| Sports (LLM) | 18.9 | 18.2 | 18.9 | Knowledge |
| COVID-19 (LLM) | 15.6 | 13.3 | 12.3 | Knowledge |
| Property (LLM) | 11.3 | 10.6 | 11.2 | Knowledge |
| Diarrhoea (LLM) | 9.9 | 9.3 | 12.3 | LLM |
| ForMed (LLM) | 9.0 | 8.3 | 9.6 | LLM |

Lastly, Figure 5 illustrates how the frequency of questionnaire responses marked as *Very Accurate* or *Mostly Accurate*, categorised by case study, varies with network size. There is a clear tendency for trust in these graphical structures to decrease with network size for graphs constructed by humans and LLMs. However, while a weaker trend appears for causal ML, the pattern is less distinct. This outcome could be due to larger networks providing more opportunities for errors in edges to be identified by participants, thereby lowering the accuracy scores, whereas the poorer accuracy ascribed to smaller causal ML graphs may also be due to the fact that counter-intuitive arcs - which causal ML tend to generate - may be easier to spot in smaller networks.

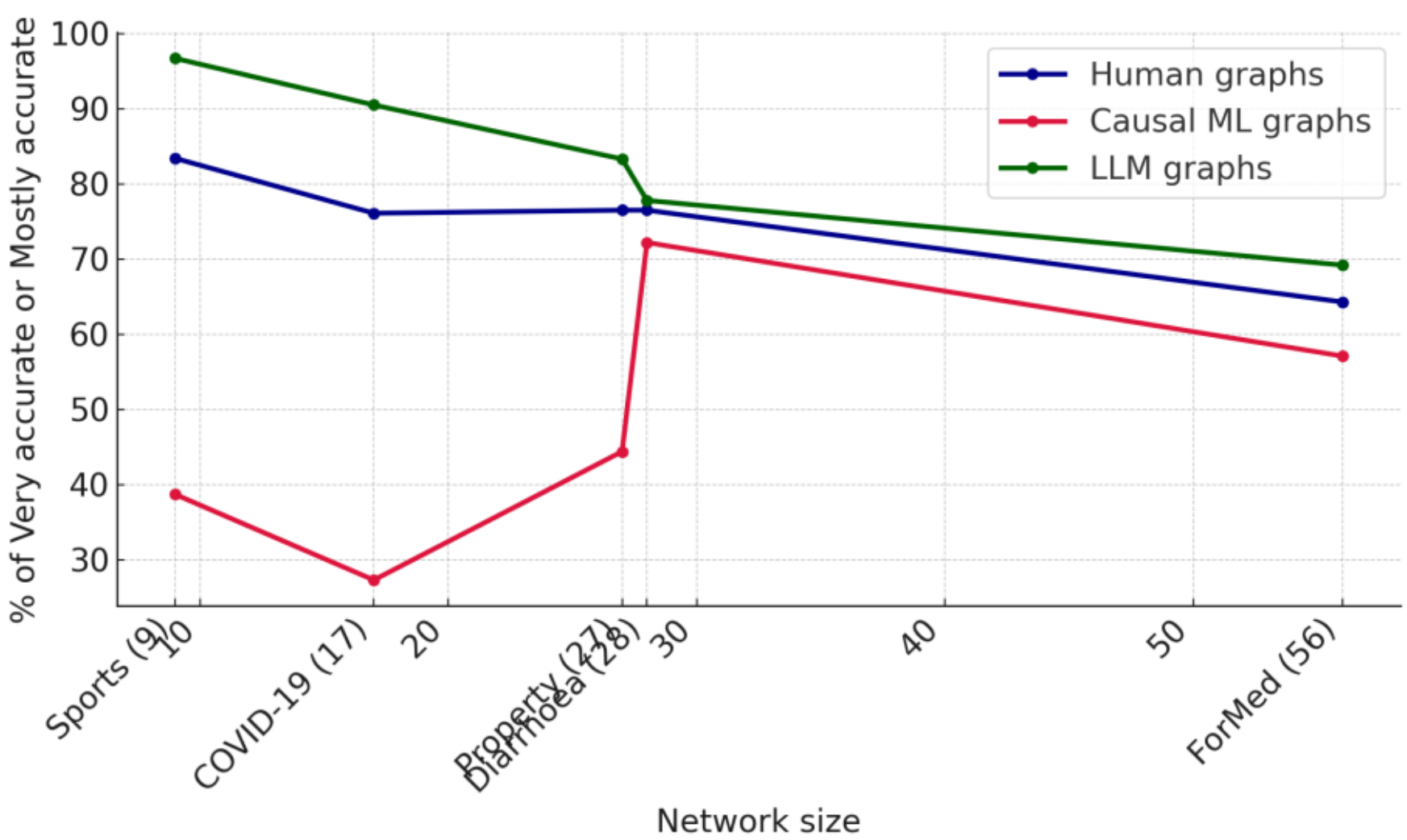

**Figure 5.** The frequency of questionnaire responses marked as *Very Accurate* or *Mostly Accurate* for relationships depicted in graphs categorised by case study, and ordered by network size.

**Table 7.** Key comments left by questionnaire participants.

| Case study | Graph | Comments |
|---|---|---|
| Sports | Knowledge | 1. "*It seems AI generated because it is* ***very symmetrical****.*"<br>2. "*The* ***symmetric nature of the graph*** *and the absence of clearly counter-intuitive arcs led me to believe this was mostly likely created by a human.*"<br>3. "*This looks like a human produced graph. A human would think of two teams in a football match as having the same variables but with different values. Humans would* ***emphasize symmetry*** *of the graph as a result.*"<br>4. "*I don't think my knowledge level gives me enough confidence to say this is very accurate when (despite being someone who used to work in Sports media) but it looks pretty conducive. If i was being critical of my own assessment, I'd say my opinions has been based on the fact that this is* ***laid out symmetrically****, in an intuitive way. Hence i think a human designed it.*" |
| Sports | Causal ML | 1. "*If this modelling is for the match simulation, RDlevel should not be the end/target node, but HDA should be the end node.* ***Human can and LLM would understand what to achieve is match result, and HDA is the end node from the context****. (but this is the case only if LLM is given a well-instructed prompt they can understand what to do)*"<br>2. "*I judged most of the relationships to be correct, except that* ***team rating was an effect of possession and natch result rather than a cause of these*** *which is what I would expect human/LLM to say … hence why I thought this most likely to be created by Causal ML*"<br>3. "*Obviously it's not human knowledge based and I think* ***any constraint-based algorithm would have got it more accurate****, so it's probably an LLM result. (I don't have enough knowledge about LLM)*"<br>4. "*This looks like a graph produced by a Causal ML.* ***Noise in the data or latent variables frequently cause the model to reverse connections****, such as possession -> RDlevel instead of RDlevel -> possession.*"<br>5. "*The only inclination i get that this might be done by a human is because of the positioning of RDlevel. On the one hand, this is a rating, perhaps it becomes an arbitrary measure in causal relationships, and is actually only a reflection of the state of the game. On the other hand, perhaps because this reflects the state of the game, it can be understood as a casual variable. I don't know. However, my conclusion is that this is perhaps a mistake by an LLM. Also,* ***the chain of cause is too simplistic****, I think possession proportion would influence the number of shots on target directly (even if this is through an implicit relationship e.g. possession increases the number of shots taken which equates to increases in shot accuracy and therefore shots on target.)*" |
| Sports | LLM | 1. "*If it was drawn by human knowledge it should have had the edge from RDlevel to possession probably!*"<br>2. "*This looks like an LLM produced graph.* ***They tend to get "most" but "not all" of the connections right****. Tell tale sign is the lack of RDlevel -> Possession connection a human would make from the start.*"<br>3. "*All* ***causal directions seemed correct but with some missing*** *making me think LLM the most likely creator*" |
| COVID-19 | Knowledge | 1. "*Again most of* ***causal relationships seemed corect, but less comprehensive*** *than Graph 2 making me think it might be more likely produced by an LLM*"<br>2. "*This looks like an LLM produced graph.* ***LLMs tend to get most but not all of the connections right****. However, they make reasoning errors, such as Deaths with Covid on Certificate -> Second Dose Uptake. A human would probably think in terms of reduction and say Second Dose Uptake -> Deaths with Covid on Certificate.*"<br>3. "*I think that* ***Graph #3 is the worst one here*** *(for example, I do not understand the meaning of the connection Deaths_with_COVID_on_certificate => Second_dose_uptake). A human could not produce this graph.*" |
| COVID-19 | Causal ML | 1. "*I think work and school activity is more likely to cause transportation activity. There should probably be a link from new/re infections to hospital admissions. Patients in MVB more likely to cause deaths with covid on certificate.*"<br>2. "*death with Covid on certificate and MVB direction, transportation activity and lockdown* ***direction does not make sense****.*"<br>3. "*This seemed to have a number of* ***counter intuitive relationships*** *e.g. Deaths by Covid --> Persons in MVBs typically produced by Causal ML*" |
| | | 4. *"This looks like a graph produced by a Causal ML. Tell tale* ***signs are reversals****, such as Positive Test -> New Infections instead of New Infections -> Positive Test, and the sparse connections, probably due to the model not being able to find the right connections or minimizing the number of connections."*<br>5. *"Looks* ***to simple to be the product of an algorithm*** *- looks like it has been built out with domain knowledge."*<br>6. *"I suggest that* ***correct connections are mostly missing here****."* |
| COVID-19 | LLM | 1. *"the position of facemask and direction of reinfection->positive_test->new-infection is not convincing."*<br>2. *"This had a set of* ***plausible sets of cause and effect****, with for instance, a comprehensice set of causes for New infections."*<br>3. *"This looks like human produced graph. Humans tend to* ***have a target variable in mind and build the graph around it****. In this case, the target variable is the number of new infections, which has a huge number of incoming connections."* |
| Property | Knowledge | 1. *"Relationships seemed* ***mostly correct but and 'well-structured'*** *making me think this was most likely human-generated"*<br>2. *"This looks like an LLM produced graph.* ***Most connections are correct, but there are some errors****, such as Income Tax -> Rental Net Profit Before Interest."* |
| Property | Causal ML | 1. *"This seemed to be* ***missing key relationships****"*<br>2. *"This looks like a graph produced by a Causal ML. Tell tale sign is the* ***sparse connections****, probably due to the model not being able to find the right connections or minimizing the number of connections."* |
| Property | LLM | 1. *"This seemed to have the* ***most comprehensive range of cause and effects****, which seemed plausible that it might be created by an LLM"*<br>2. *"This looks like human produced graph. Humans tend to have* ***a target variable in mind and build the graph around it****. In this case, the target variable is the net profit, which has a huge number of incoming connections."* |
| Diarrhoea | Knowledge | 1. *"This seemed* ***the most "well-structured" graph*** *making me think a human was the most likely creator"*<br>2. *"This looks like human produced graph. There is a* ***clear tiered structure between the variables*** *showing a hierarchy of importance. And one variable is centralized as the cause of most of the other variables (Economic Wealth Quintile)."*<br>3. *"there are* ***more dependencies in this model in general****, i think this makes it more likely to be produced by an algorithm or LLM."* |
| Diarrhoea | Causal ML | 1. *"E.g.* ***watching tv can't cause the mother's education****, i think region more likely to affect language than other way round."*<br>2. *"The relationship around immunisation and vitamin A1 and the direction of region and language group seem* ***not correct****. Also, the cause of the diarrhea are only breast and bottle feeding and there should be more factors cause the diarrhea"*<br>3. *"Most relationships seemed plausible, but* ***some seemed wrong*** *e.g. immunisation -> EarlyBreastFeeding because this is in the wrong temporal order."*<br>4. *"This looks like a graph produced by a Causal ML. Tell tale sign is the* ***reversal of connections****, such as CUL Language Group -> GEO Region, which a human would probably think of as GEO Region -> CUL Language Group."* |
| Diarrhoea | LLM | 1. *"Most* ***relationships seemed plausible, but therew ere too many*** *making me think LLM was the most likely the creator."*<br>2. *"This looks like an LLM produced graph. Most connections are correct, but there are* ***some missing connections****. LLMs usually need a few rounds of prompting to get all the possible connections out of them."*<br>3. *"This graph seems a lot* ***more disjointed than the others. It's less interconnected****, with features being introduced at all tiers of the graph."* |
| ForMed | Knowledge | 1. *"Graph seemed to have* ***implausible causal relationships*** *(e.g. Age is a cause of Violence) making me think this was most likely created by Causal ML"*<br>2. *"This looks like an LLM produced graph. Most connections are correct, but there are* ***some missing connections*** *and reversal which a human would probably not make."* |
| ForMed | Causal ML | 1. *"Seemed to have some* ***counter-intuitive relationships*** *e.g. CannabisUse was a cause of Age making me think Causal ML most likely creator"*<br>2. *"This looks like a graph produced by a Causal ML. Tell tale signs are the reversal of some connections and the* ***sparse connections****."* |
| ForMed | LLM | 1. *"Most relationships seemed correct, but* ***graph rather dense*** *(e.g. many many direct causes of Violence) making me think LLM might be the creator."*<br>2. *"This looks like human produced graph. Humans tend to have a* ***target variable in mind and build the graph around it****. In this case, the target variable is the Violence, which has a huge number of incoming connections."* |

### *3.2. Using GPT-4 to guide causal ML*

As described in Section 2.3, we also test the usefulness of GPT-4 in terms of using its output as causal constraints to restrict or guide the search space of graphs explored by causal ML algorithms. The results presented here are based on four case studies involving real (not synthetic) datasets, as shown in Table 3. These results consider three different confidence levels of constraints, also described in Table 3, and three types of constraints: required edges, temporal order, and initial graph, as described in Section 2.3. Additionally, eight algorithms from different classes of learning, which support some or all of these types of constraints, are utilised as detailed in Table 4.

Figure 6 presents the overall impact of GPT-4 constraints on structure learning. Specifically, the results measure the relative impact on the graphical structures learnt by the causal ML algorithms with real data, comparing scenarios with and without GPT-4 constraints, and with reference to the knowledge graph for each case study as determined by domain experts. Each sub-chart summarises the results using the different metrics of F1, BSF, SHD, and BIC scores, for each rate and type of constraint across all algorithms and case studies.

The F1, BSF, and SHD scores represent graphical metrics that measure the distance between two graphical structures. With reference to the confusion matrix, the SHD score considers the *false positive* and *false negative* edges between the two graphs, the F1 score includes those considered by SHD plus the *true positive* edges, and the BSF score further includes those considered by F1 plus the *true negative* edges. Note that because SHD does not account for true positive nor true negative edges, it is known to be biased in favour of sparser graphs. However, the SHD score is widely used in the literature, and while we present the SHD scores to enable cross-comparisons between studies, most of our focus will be on the F1 and BSF metrics. Lastly, in contrast to the graphical metrics, the BIC score is a model-selection function that estimates how well the learnt model, balances between data fitting and model dimensionality.

The results presented in Figure 6 show that all three graphical metrics agree that the GPT-4 constraints help the algorithms output a graphical structure that is closer to those produced by domain experts, compared to the corresponding graphical structures learnt without GPT-4 constraints. The results also indicate that, amongst the different types of constraints, the GPT-4 constraints are most effective when employed as *required edge* constraints, irrespective of the rate of constraints. The *initial graph* constraints do generate a positive effect too, but not as strong and consistent as *required edge* constraints, whereas the *temporal* constraints produce mixed results.

For *required edge* constraints, both the F1 and BSF scores show that the results are stronger at a 33% rate of constraints, implying that the constraints are more beneficial when extracted from the set of edges that appear in at least a third of the 10 GPT-4 prompts. This goes against our initial expectation, which expected the results to be stronger at a 67% rate of constraints, where the edges constrained are restricted to those that appear in at least two-thirds of the 10 GPT-4 prompts, thereby increasing the confidence in the set of constraints due to larger agreement between GPT-4 prompts. On the other hand, we observe the reverse effect for *initial graph* constraint, with mixed effect in other cases, and so this observation does not seem to be consistent across all types of constraints and metrics.

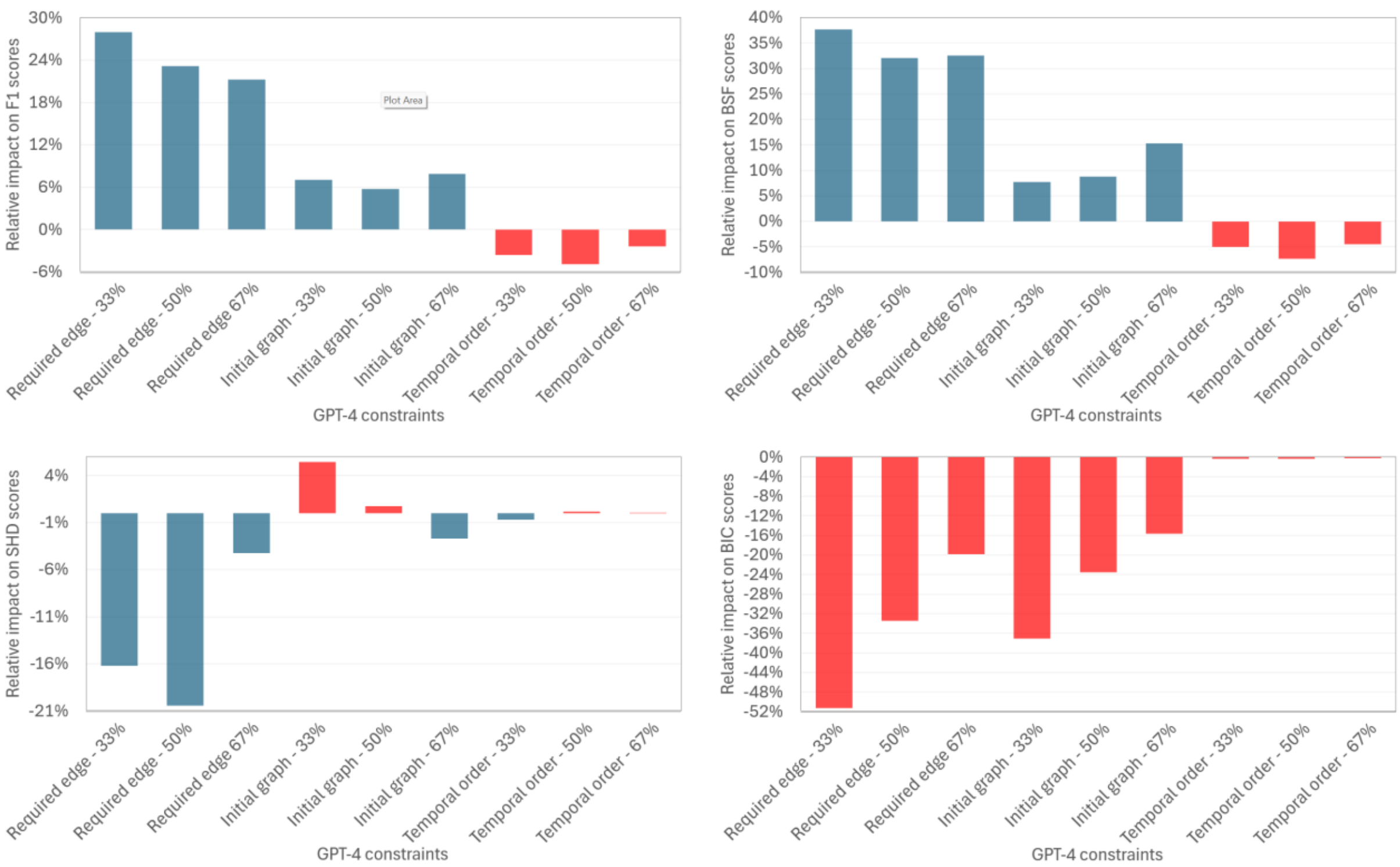

**Figure 6.** The impact of GPT-4 constraints on structure learning in terms or relative change in CPDAG score, over all algorithms and case studies, and based on the specified threshold rates about the proportion of times the same edge (constraint) appeared across each of the 10 GPT-4 prompts per case study. A lower percentage in the legend indicates a higher number of GPT-4 constraints. Blue coloured bars indicate an increase in accuracy, whereas red coloured bars indicate a decrease in accuracy. BIC scores exclude PC-Stable since PDAG outputs could not be converted into a CPDAG.

The BIC score, on the other hand, decreases across all cases. This is not necessarily surprising since the score-based algorithms are designed to find optimal or close-to-optimal structures that maximise the BIC objective score. This means that the added constraints prohibit the algorithms from exploring parts of the search space that may contain a higher objective score. For example. notice how the higher numbers of constraints, in the 33% and 50% cases, tend to decrease the BIC score faster than when the quantity of constraints is lower, as in the 67% case. While it may be counterintuitive for constraints to increase graphical scores but decrease model-selection scores, it is consistent with previous studies that show that the graphs constructed by domain experts often yield BIC scores that are distant from the optimal graph – as judged by BIC - within the search space of graphical structures. This discrepancy arises because knowledge-based graphs tend to overlook model dimensionality, and this study further highlights the weaknesses of traditional objective functions in recovering graphical structures that align with expertly-constructed causal graphs.

Figure 7 presents the range of graphical scores produced across the different structure learning settings using box-plots. Unlike Figure 6 which contradicted our initial expectations – that fewer, more 'certain' GPT-4 constraints (at the 67% threshold) would be more effective than more, less 'certain' constraints - Figure 7 partly supports this expectation. This is because it shows that the fewer constraints generated at 67% threshold effectively limit the number of low graphical scores. However, these fewer constraints do not lead to the higher graphical scores observed at 33% and 50% thresholds, explaining why Figure 6 supports these lower thresholds. Overall, the 67% threshold seems to reduce the variability of the results, effectively avoiding the lowest scores but also failing to reach the highest scores.

Figure 8 provides a detailed analysis of the impact of GPT-4 constraints on F1 scores at the individual algorithmic level, based on the aggregated results shown in Figure 7. The findings indicate that the trends observed in Figure 7 largely extend to each individual algorithm, reinforcing confidence in the positive effect of GPT-4 constraints on structure learning, especially when the constraints are applied as *Required edges* or *Initial graph*.

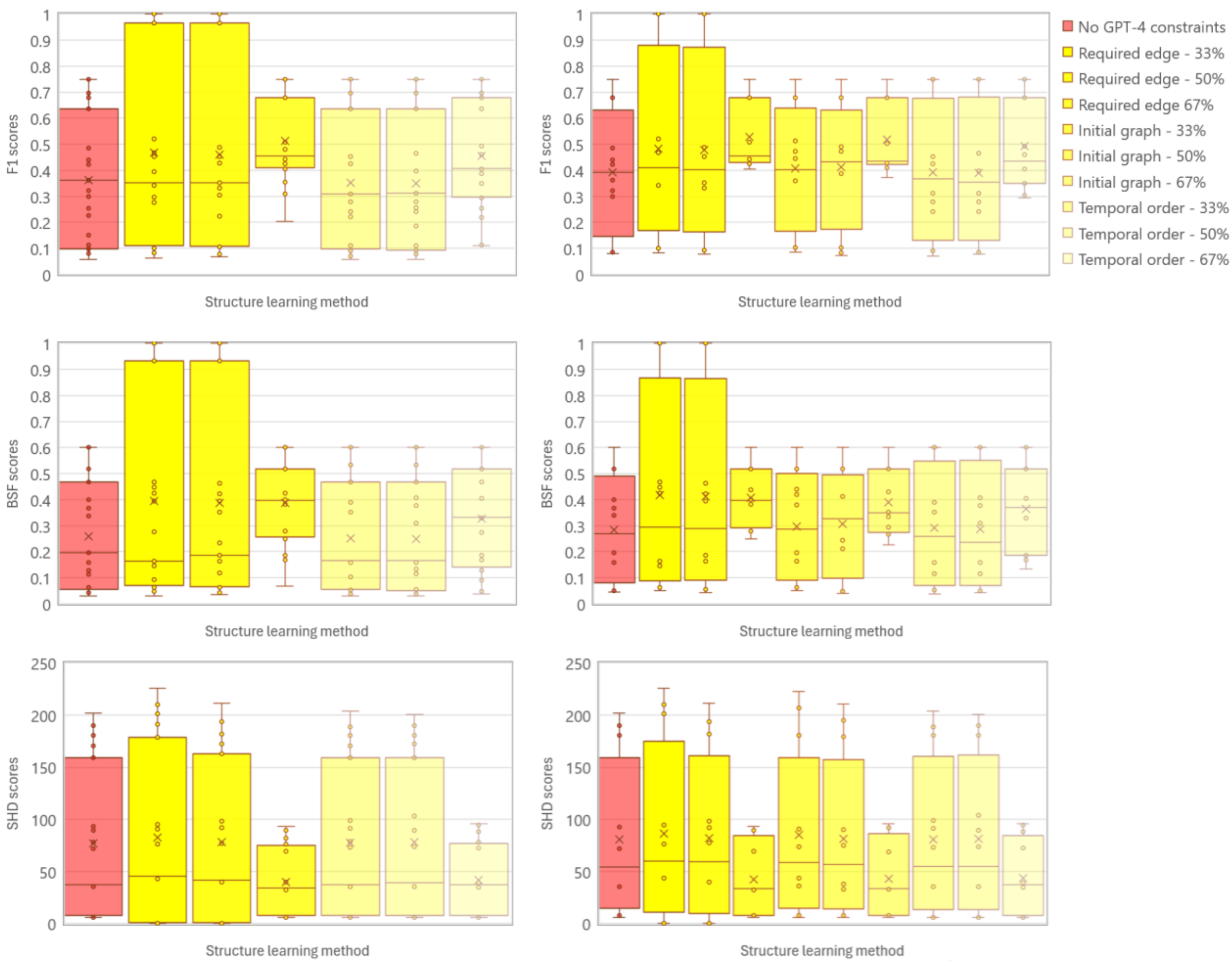

**Figure 7.** Box-plots on the comparison between the specified graphical metric scores produced by causal ML without GPT-4 constraints (in red), and causal ML restricted or guided by different types and rates of GPT-4 constraints (in various shades of yellow), where each box illustrates the interquartile range, the horizontal line inside each box is the median, x is the mean, *o* are the inner values that fall within the range from the lower quartile (Q1) to the upper quartile (Q3), and the whiskers represent the minimum and maximum values. A lower percentage in the legend indicates a higher number of GPT-4 constraints. The charts on the left include all causal ML algorithms considered, but do not present the *Initial graph* type of constraint (to avoid bias) since it was not supported by all algorithms. The charts on the right summarise the results across all types of constraints, restricted to the algorithms that support all of them (HC, TABU, MAHC, and SaiyanH).

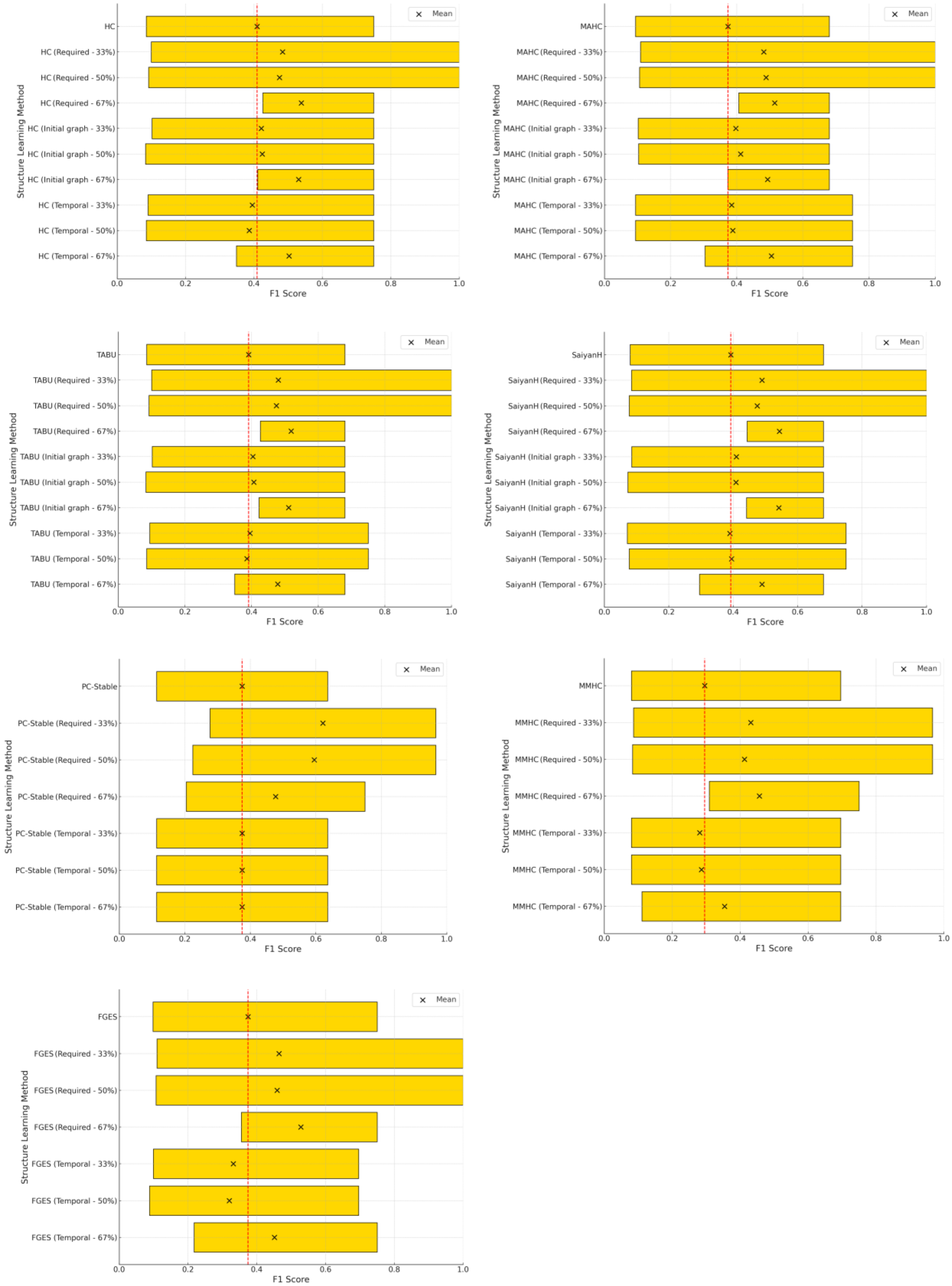

**Figure 8.** An extended comparison of F1 scores for each structure learning algorithm, based on the aggregated results shown in Figure 7, both with and without GPT-4 constraints. Each bar represents the range of F1 scores across learnt and knowledge-based networks, ranging from the minimum to the maximum F1 score , with the mean indicated by an 'x'. A red vertical dashed line highlights the mean F1 score for the top experiment which represents learning without GPT-4 constraints, providing a reference for comparison with the other experiments under GPT-4 constraints.

Lastly, Figure 9 illustrates how the average change in F1 score varies across case studies, ordered by network size, similarly to Figure 5 which presents questionnaire responses. While Figure 5 shows a clear relationship between questionnaire responses and network size, Figure 9 does not demonstrate a similar association, suggesting a weak relationship between network size and the impact of GPT-4 constraints. Figure 9 also highlights cases where GPT-4 constraints were not beneficial to causal ML; i.e., a) the *Required edge* constraints at 33%, which had minor positive impact on the Sports case study (0.2% increases in overall F1 score), and b) the *Temporal order* constraints at 33% and 50%, which led to small negative impacts on the COVID-19 case study (-3% in overall F1 score) and the Diarrhoea case study (-2.1% in overall F1 score).

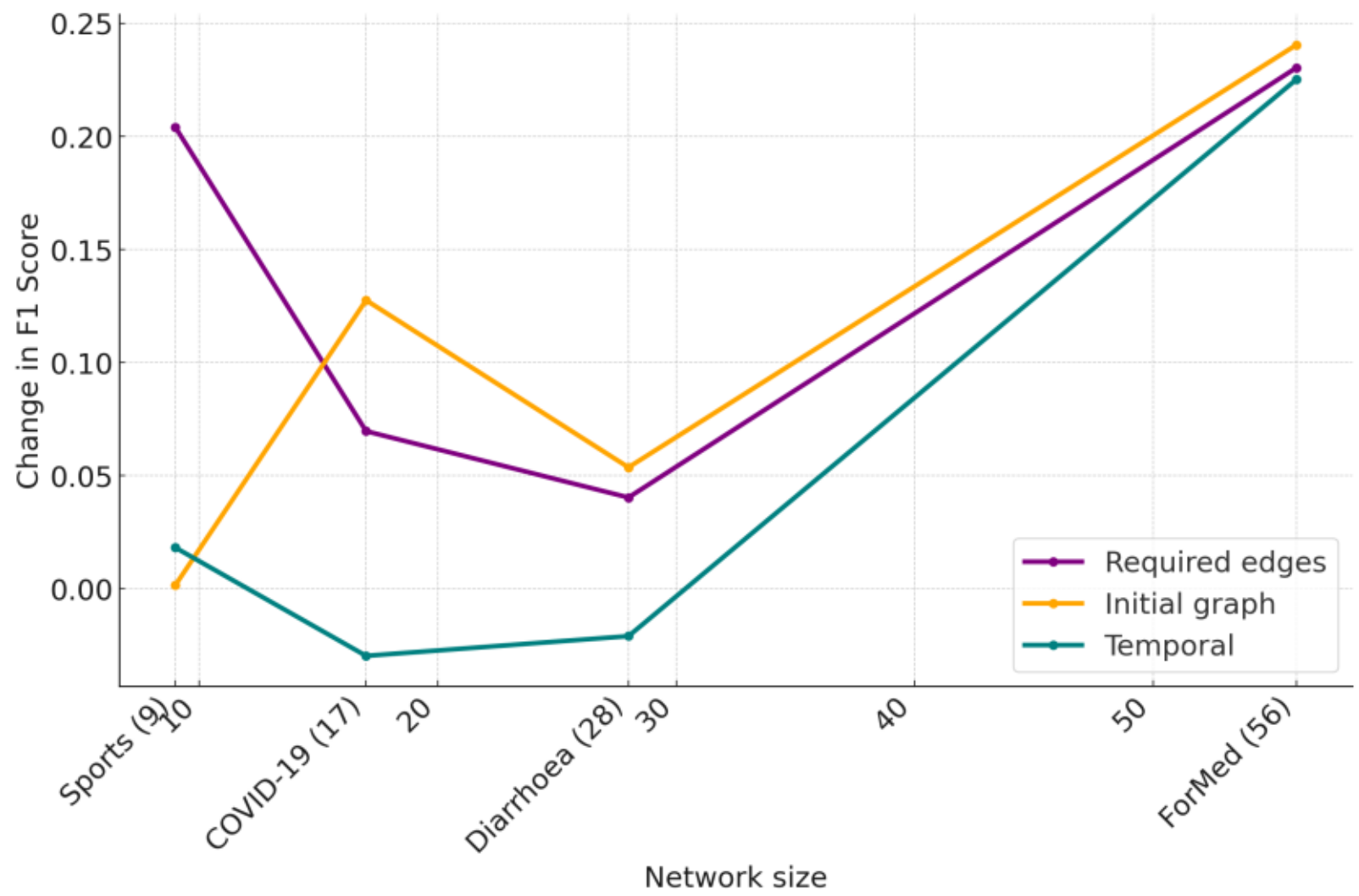

**Figure 9.** The average change in F1 score, categorised by the type of GPT-4 constraint type and by case study, ordered by network size.

## 4. Discussion and concluding remarks

LLMs transform data and user input into numerical representations known as tokens. These tokens capture the sematic meaning of the words, and the trained models appear to understand context through layers of neural-network transformations. This process helps LLMs generate coherent and relevant response. Therefore, while LLMs are not designed to disentangle correlation from causation, they often produce output that appears to be causally valid due to their ability to recognise sophisticated patterns. This can create the impression that the models understand causality, but it is important to highlight that their apparent causal reasoning is a byproduct of their training process rather than a true comprehension of causal relationships.

Still, because the output of LLMs is now perceived to be much more causally valid than we would expect from an associational model, the role of causality in LLMs is becoming an area of significant debate. This study adds to this emerging field by exploring the usefulness of GPT-4 outputs in terms of causal reasoning, and comparing them to those derived from domain experts and those learnt from data using causal ML algorithms.

It is important to clarify that the aim of this study is not to evaluate the performance of GPT-4 based on the accuracy of the causal relationships it generates for each case study, as we cannot definitively know if these relationships are correct.

Instead, the objective is to assess GPT-4 in terms of a) the impact these relationships have when used as constraints in the structure-learning process of causal ML algorithms, and b) how participants perceive the causal relationships generated by GPT-4 compared to those produced by causal ML models and domain experts. The contributions of this paper are two-fold; it demonstrates that:

a. questionnaire participants find it difficult to distinguish between graphs generated by GPT-4 and those by domain experts, while easily differentiating these from causal ML graphs, and
b. LLMs, even when given only variable labels and no additional human-provided context, improve causal ML performance, both at the aggregate level and consistently across individual algorithms, by guiding structure-learning algorithms toward causal structures that align more closely with graphs constructed by domain experts.

We first designed a questionnaire that asked participants to predict whether a presented graph was drawn by causal ML, LLM, or domain experts, and to judge the causal accuracy of the graph. The results (refer to Table 6) show that participants correctly identified causal ML graphs, but misclassified some LLM graphs as knowledge graphs and vice versa. Causal ML graphs were the easiest to classify, and this observation is attributed to counterintuitive edges that we would not expect a domain expert nor an LLM to produce (refer to Table 7). Moreover, participants consistently rated LLM graphs as being fairly more accurate than knowledge graphs elicited from domain experts, and much more accurate than causal ML graphs (refer to Table 5).

GPT-4 has shown to be able to generate outputs for the case studies tested that are indistinguishable from, and often were judged by questionnaire participants as being more accurate than, those from domain experts. This might be because LLMs effectively summarise targeted human knowledge from sources that are assumed to be credible. This suggests that LLM outputs are likely to be valid, generating responses that are, or appear to be, well-informed. While some case studies tested in this paper might be part of the training data of GPT-4, this cannot be confirmed. This is because the specifics of the training data remain proprietary and undisclosed by the developer OpenAI; i.e., OpenAI has not released information on the data or sources used for training GPT-4. As a result, it is not possible to determine whether particular case studies or datasets were part of the model's training data or whether it has prior exposure to these specific examples. Regardless, this is not expected to impact performance, as LLMs like GPT-4 produce well-generalised outputs with an element of randomness from vast amounts of related examples and hence, it is not unreasonable to assume that removing a single example from its training process is unlikely to lead to significant changes in its output.

We also tested the usefulness of GPT-4 in terms of using its output as causal constraints to restrict the search space of graphs explored by causal ML algorithms. Through an extensive set of empirical experiments involving multiple case studies, causal ML algorithms, types of constraints, and quantities of constraints, the results show that GPT-4 consistently helps causal ML to produce graphical structures that are closer to those produced by domain experts, compared to the corresponding graphical structures learnt without GPT-4 constraints, and this result is consistent across all algorithms tested at the individual level.

Overall, our findings suggest that even though GPT-4 is not explicitly designed to reason causally, it can still be a valuable tool for causal representation. This is despite the fact that GPT-4 was provided with no domain context by humans; it was given just a set of variable labels and asked to connect them causally. Note that the variable labels are meaningful to LLMs, but meaningless to causal ML since it learns from data in an unsupervised manner. Therefore, these results potentially highlight the lowest possible performance one could expect from GPT-4 in terms of causal reasoning. Nonetheless, the results of this study suggest that GPT-4 potentially enhances current solutions for causal discovery. Despite these positive findings in favour of LLMs, and somewhat negative ones for causal ML, the latter is expected to be more effective in tackling previously unexplored problems where LLMs may struggle to generalise effectively.

This study comes with some limitations that could inform directions for future research. Firstly, the questionnaire results, though showing reasonably clear patterns, are based on a limited sample size of 32 responses, which may not be sufficient for drawing strong conclusions about human perceptions of the causal graphs. This limited participation can be partly attributed to the lack of compensation and the questionnaire's length, which required approximately 30 minutes for those who completed it in full. Secondly, the empirical experiments are restricted to case studies of small to moderate complexity, containing up to 56 variables. This limitation was necessary for the networks incorporated into the questionnaire to be readable and understandable to participants, and for the GPT-4 prompts and outputs to handle the number of variables reasonably well. Therefore, the results presented in this paper may or may not be representative of more complex real-world scenarios, such as gene regulatory networks which are not as well-understood as the case studies investigated in this paper, and where causal ML could perform better than experts or LLMs, or other high-dimensional systems in which the number of causal variables and edges is significantly larger. Thirdly, the results presented in this paper are based solely on GPT-4 and may not generalise to other LLMs not examined in this study.

## Appendix A: Questionnaire sample

Section 3 of 7

**Case study 1: Football match simulation**

**Variable descriptions (optional read)**

| Variable name | Description |
|---|---|
| RDlevel | Rating (i.e., team strength) difference between the two teams. |
| possession | Duration of the match spent in possession of the ball. |
| ATshots | Shots by the away team. |
| HTshots | Shots by the home team. |
| ATshotsOnTarget | Shots on target by the away team. |
| HTshotsOnTarget | Shots on target by the home team. |
| ATgoals | Goals scored by the away team. |
| HTgoals | Goals scored by the home team. |
| HDA | Match outcome; home win, draw, or away win. |

**Graph #1**

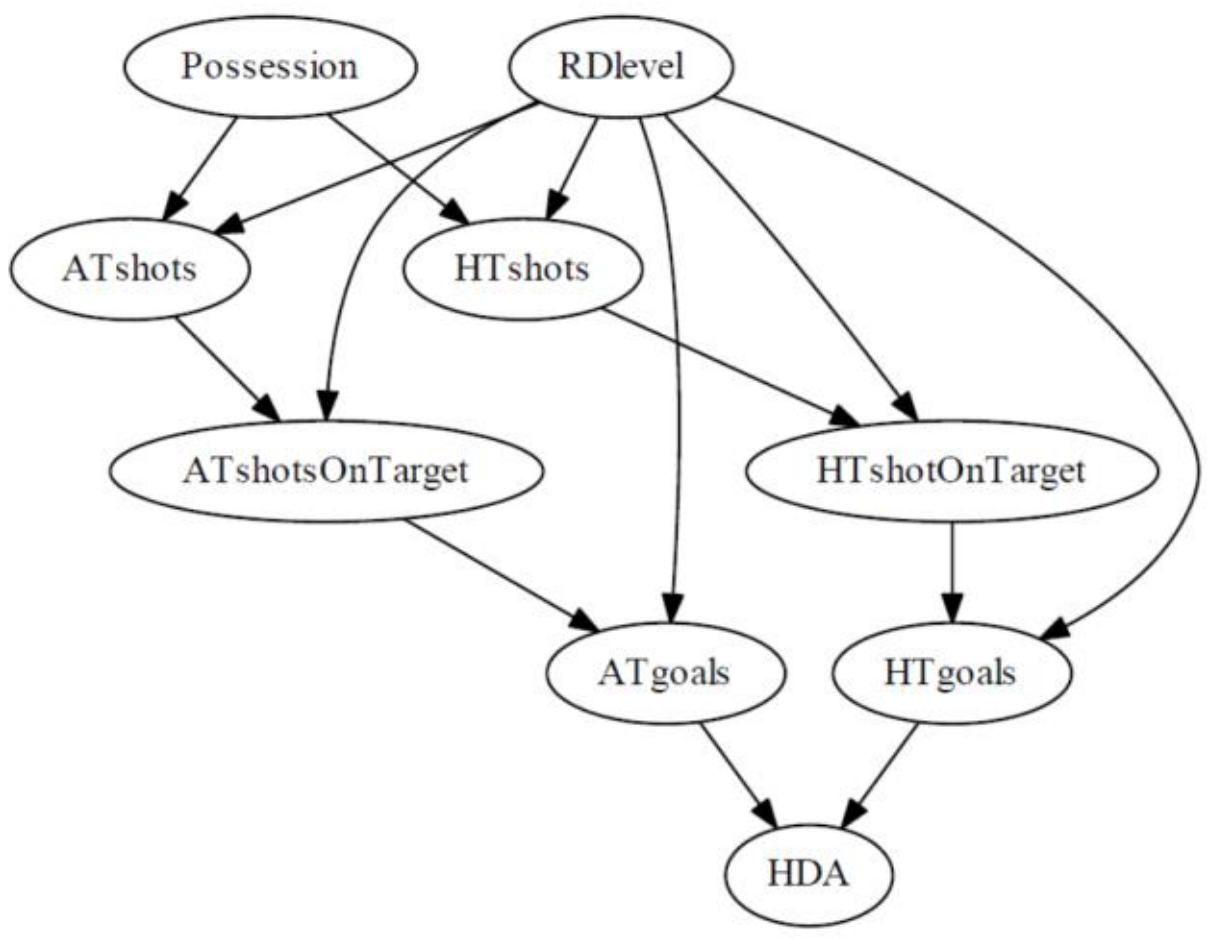

**How likely is it that Graph #1 was produced by a human, causal Machine Learning (ML), or Large Language Model (LLM)?**

| | Highly likely | Likely | Unlikely | Highly unlikely |
|---|---|---|---|---|
| Human | ○ | ○ | ○ | ○ |
| Causal ML | ○ | ○ | ○ | ○ |
| LLM | ○ | ○ | ○ | ○ |

**How accurate do you consider the causal relationships in Graph #1 ?**

- ○ Very accurate
- ○ Mostly accurate
- ○ Mostly inaccurate
- ○ Very inaccurate

**Comments on Graph #1 (optional)**

Long-answer text

**Fig A.1.** A sample of the questionnaire presenting the set of questions associated with the first graph (out of three) of the first case study (out of five).

## Appendix B: Supplementary results from the questionnaire responses

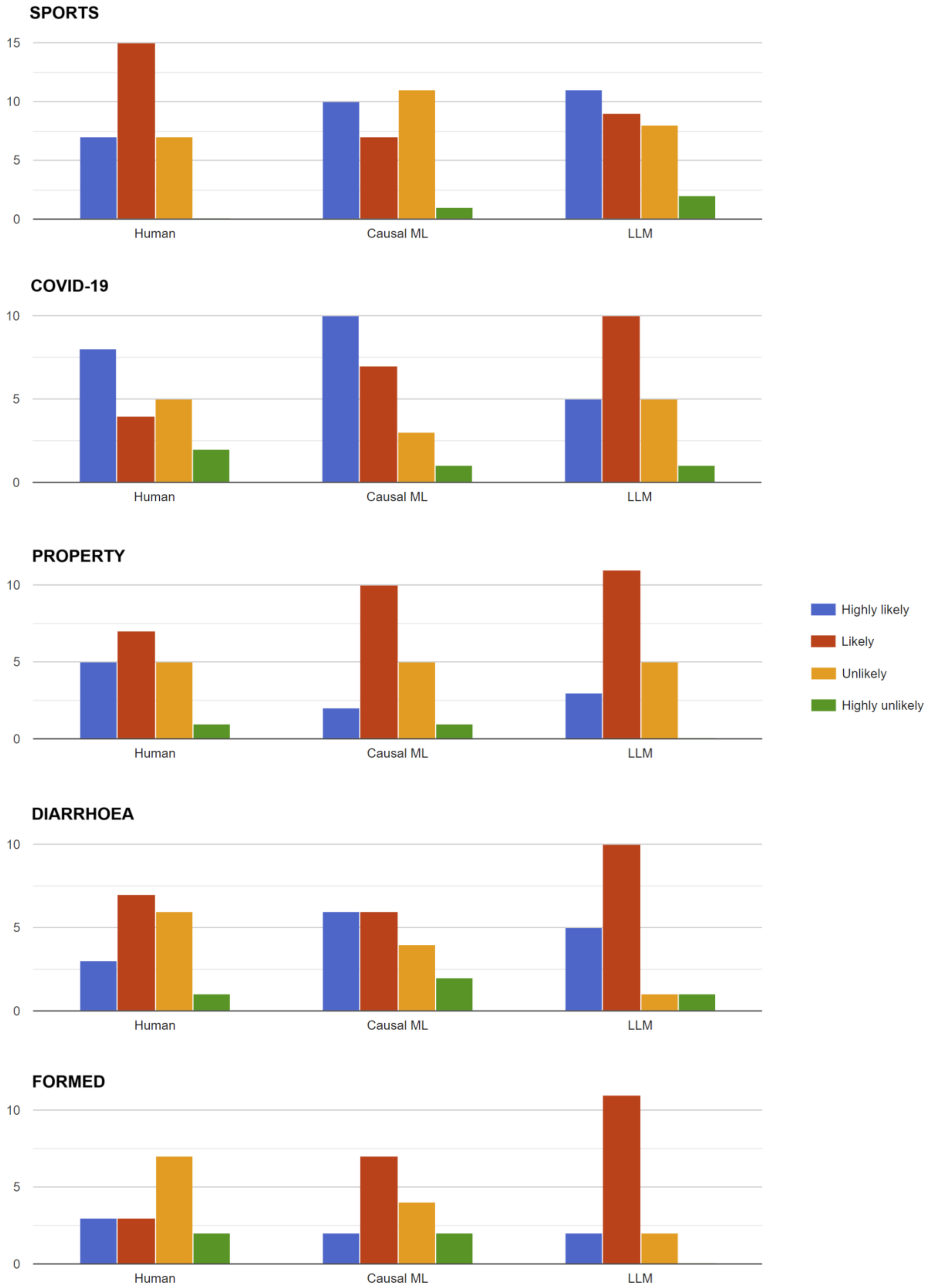

**Figure B.1.** Questionnaire responses assessing the graphical structures elicited from domain experts.

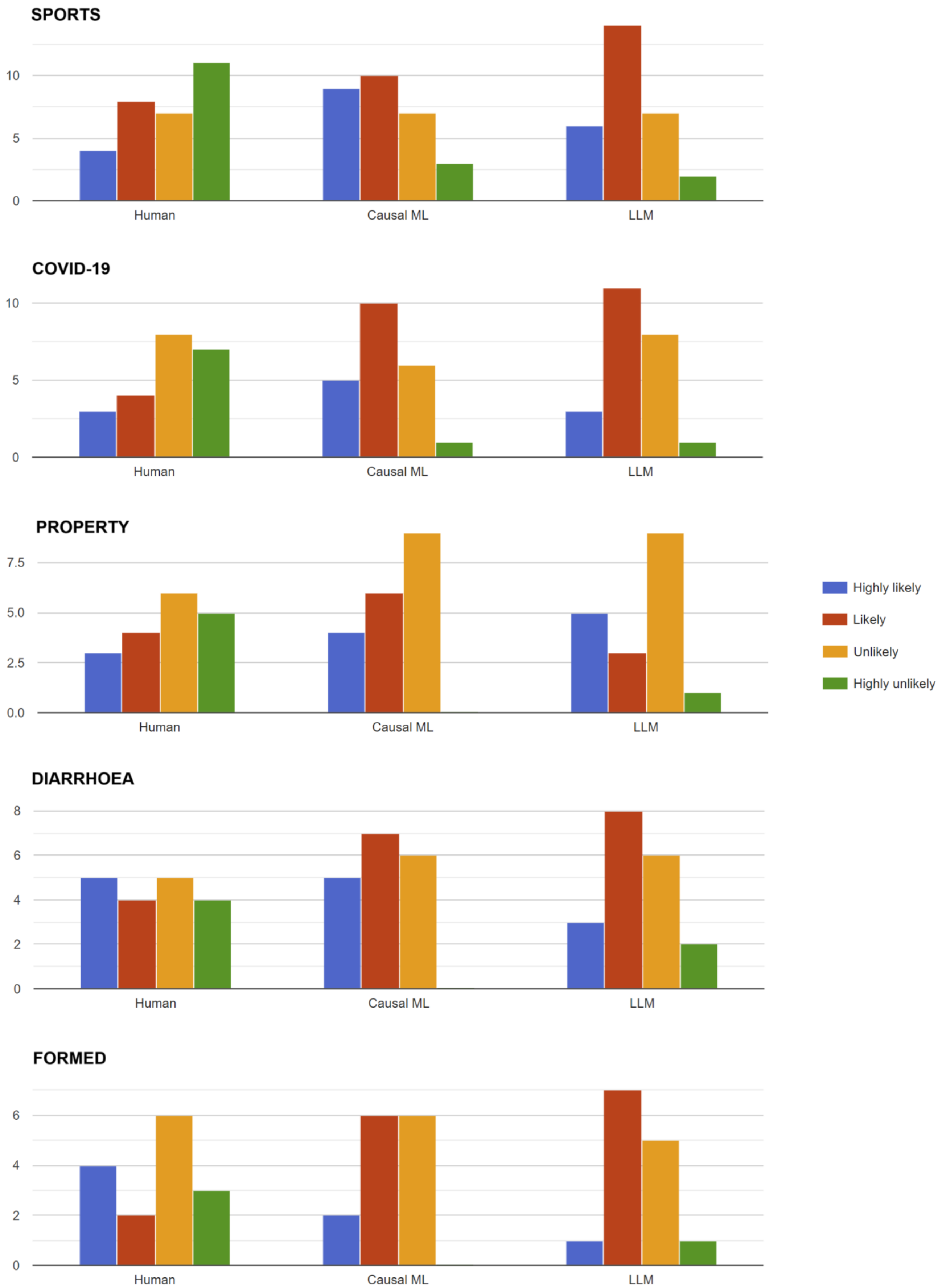

**Figure B.2.** Questionnaire responses assessing the graphical structures learnt with causal ML algorithms.

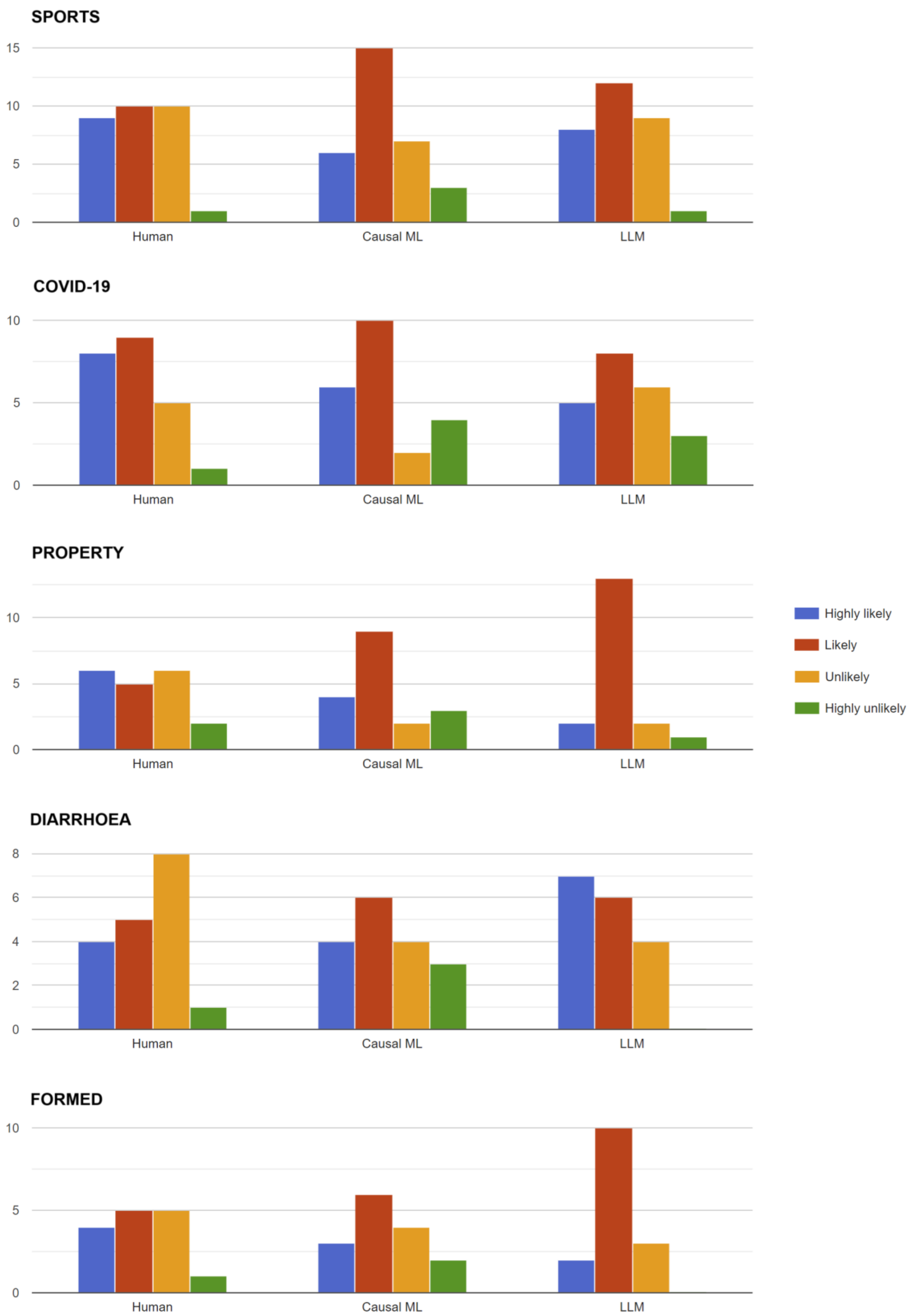

**Figure B.3.** Questionnaire responses assessing the graphical structures extracted from GPT-4.

## Appendix C: The knowledge-based, causal ML, and LLM graphical structures for each of the five case studies.

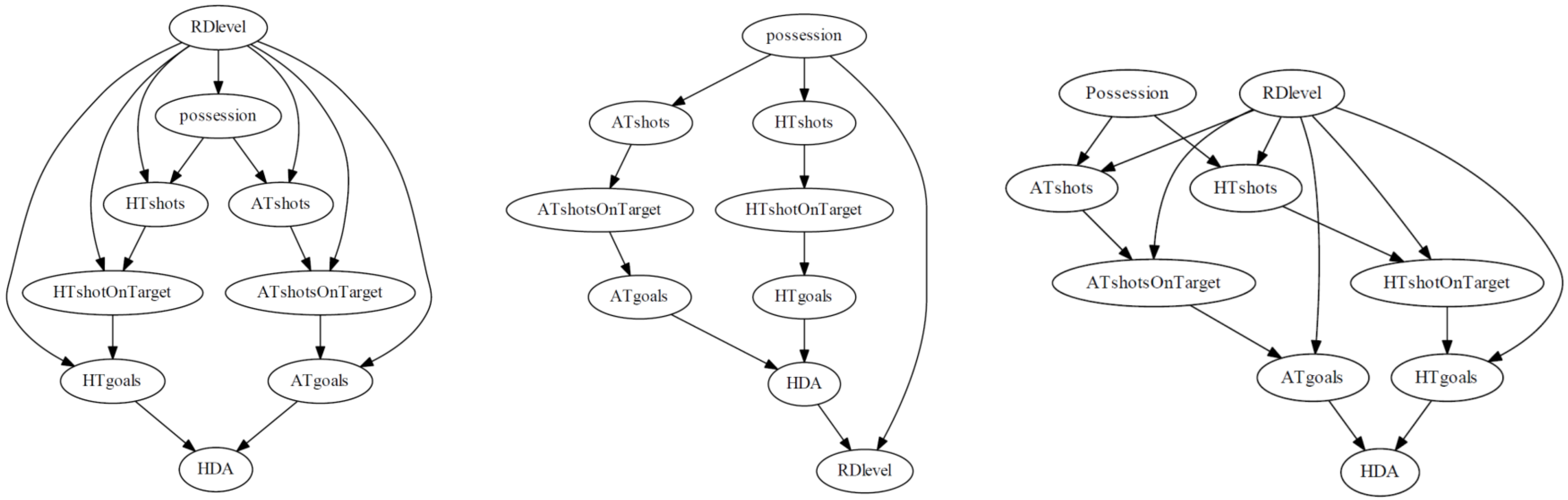

**Figure C.1.** From left to right, the knowledge, causal ML, and LLM (GPT-4) graphs for case study *Sports*.

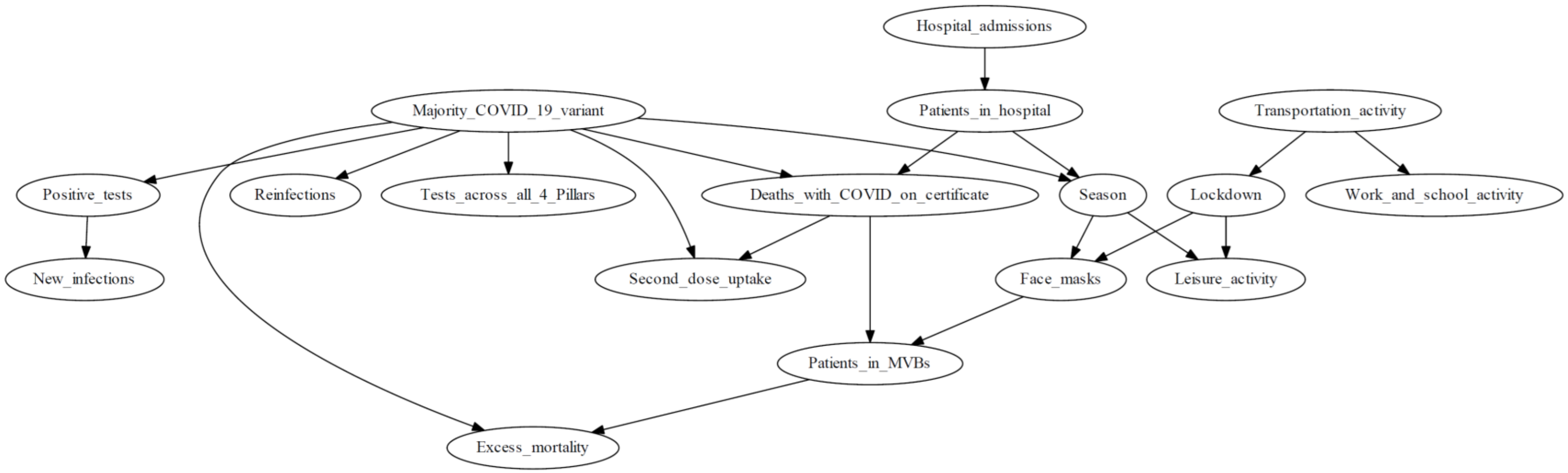

**Figure C.2.** The causal ML graph for case study *COVID-19*.

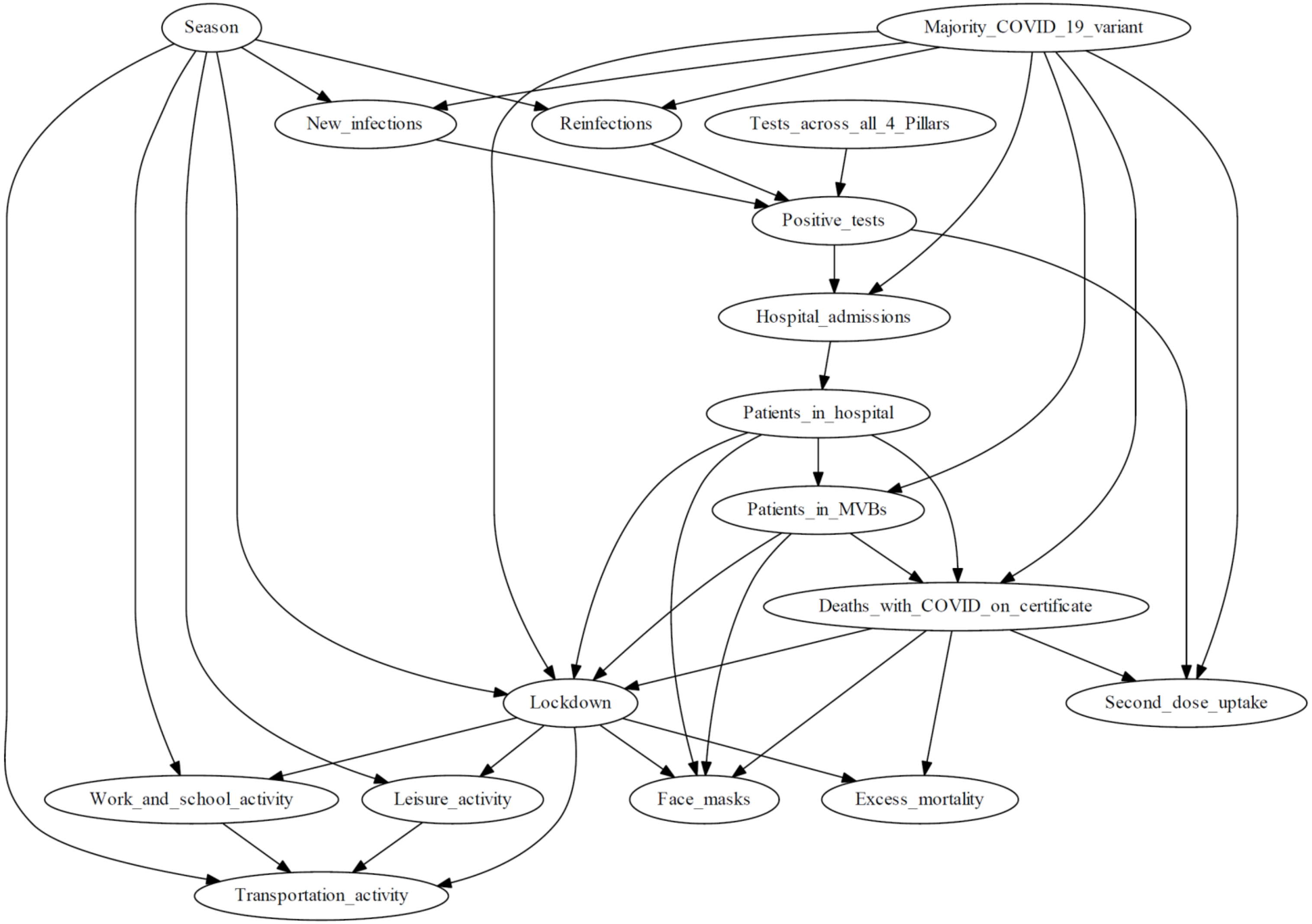

**Figure C.3.** The knowledge graph for case study *COVID-19*.

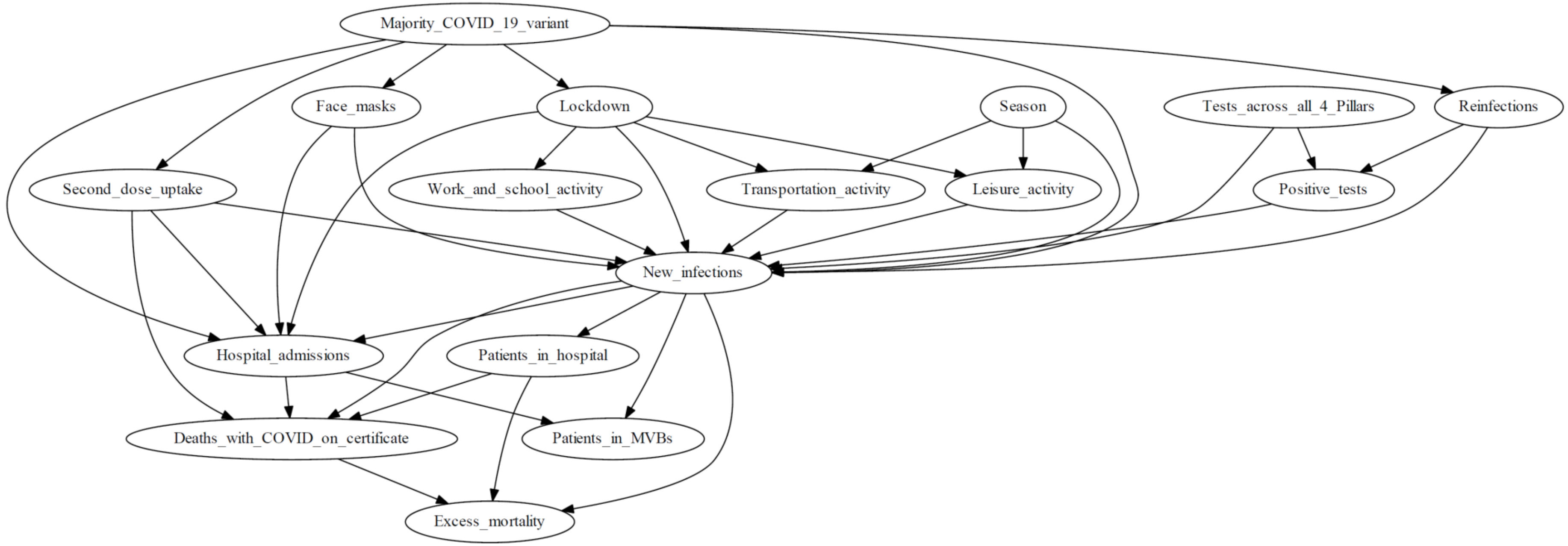

**Figure C.4.** The LLM (GPT-4) graph for case study *COVID-19*.

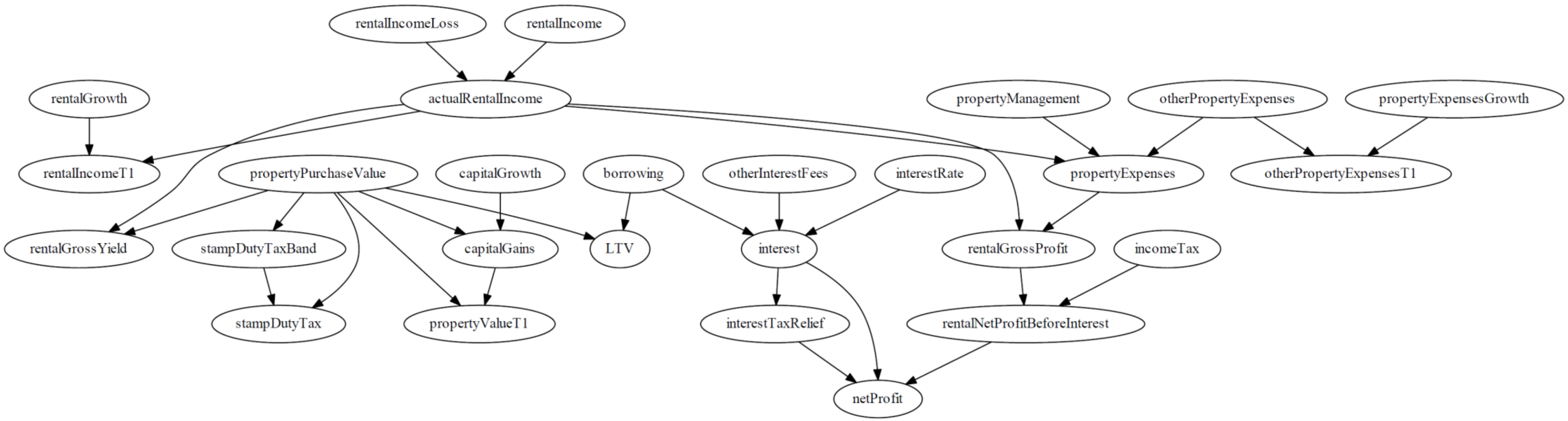

**Figure C.5.** The knowledge graph for case study *Property*.

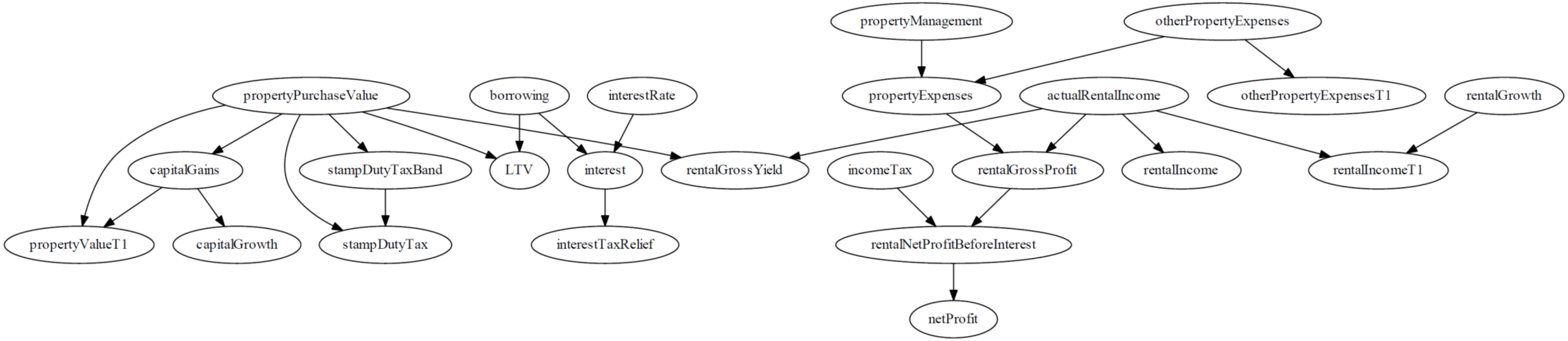

**Figure C.6.** The causal ML graph for case study *Property*.

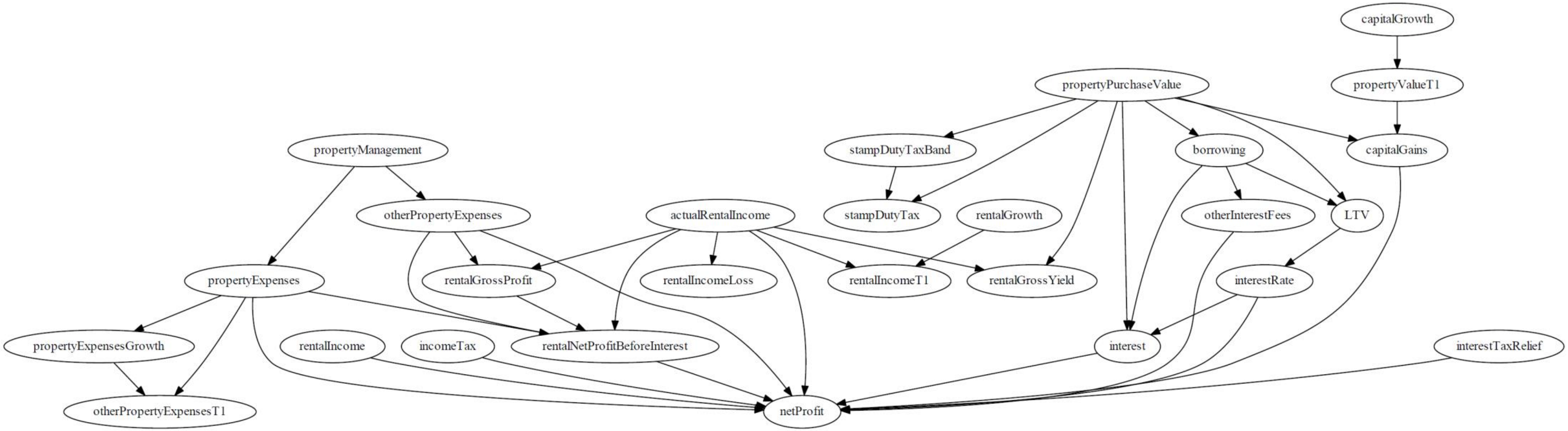

**Figure C.7.** The LLM (GPT-4) graph for case study *Property*.

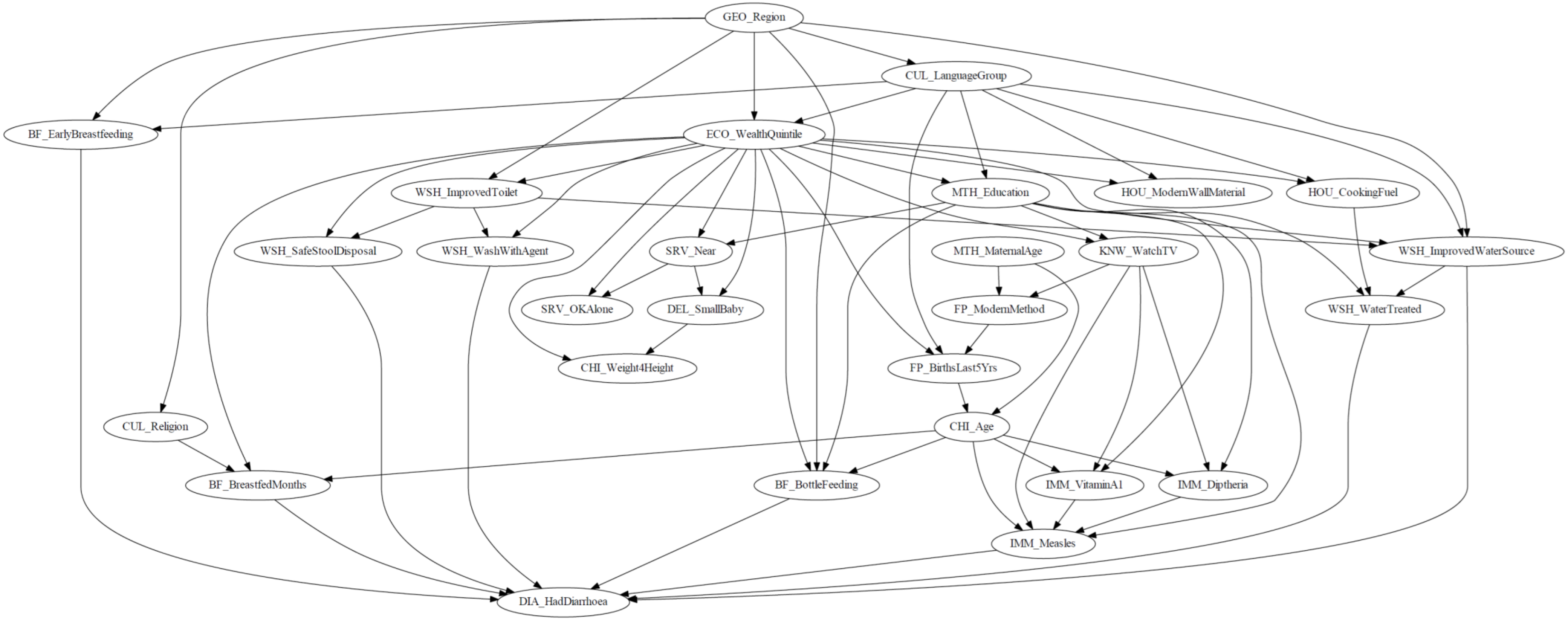

**Figure C.8.** The knowledge graph for case study *Diarrhoea*.

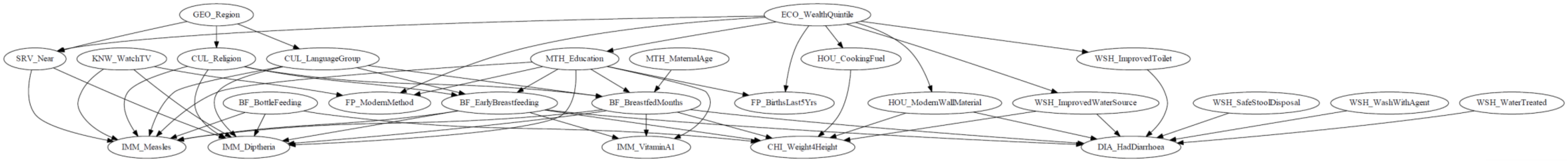

**Figure C.9.** The LLM (GPT-4) graph for case study *Diarrhoea*.

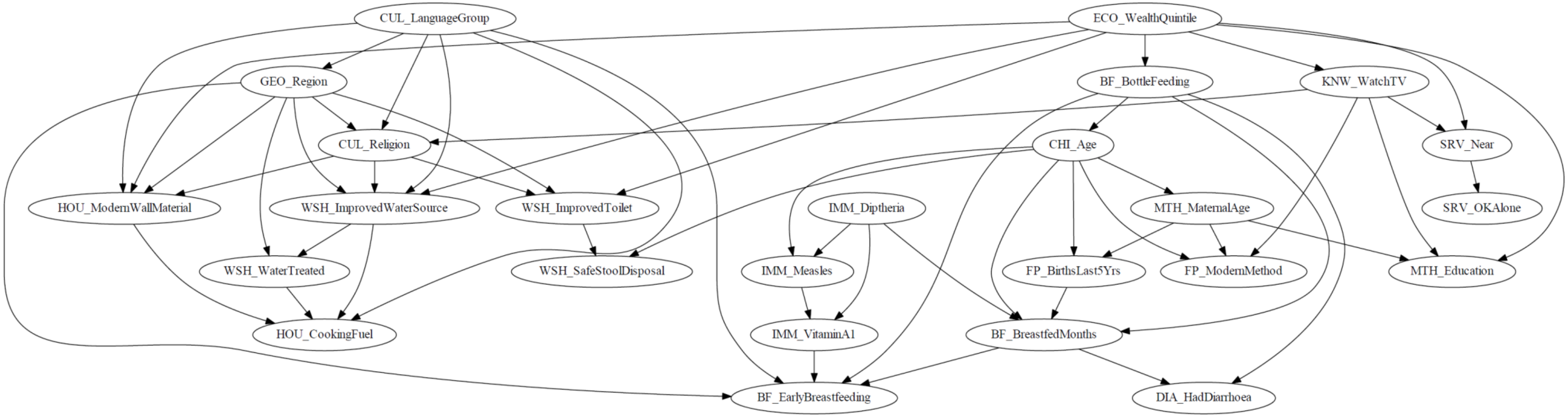

**Figure C.10.** The causal ML graph for case study *Diarrhoea*.

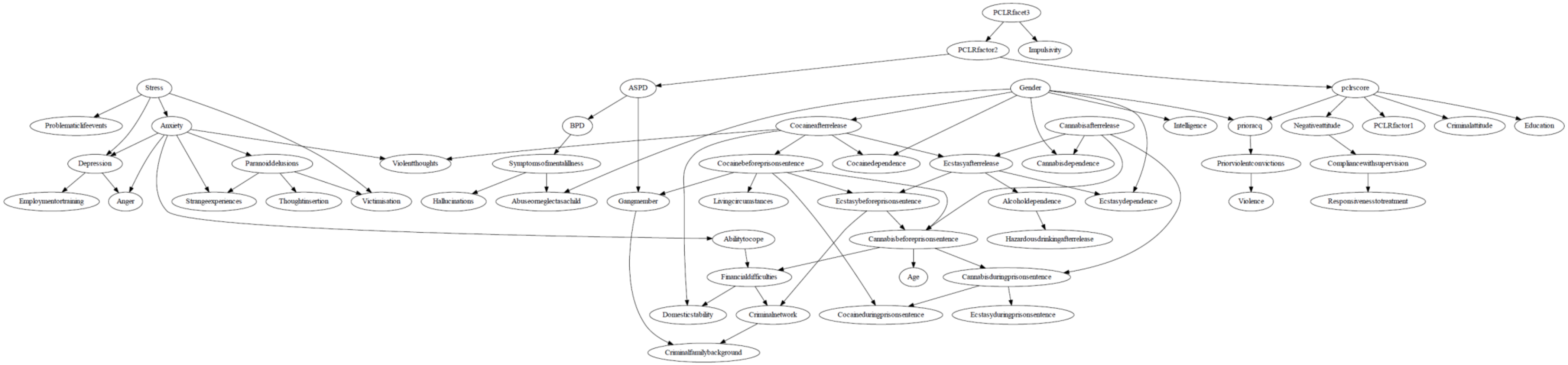

**Figure C.11.** The causal ML graph for case study *ForMed*.

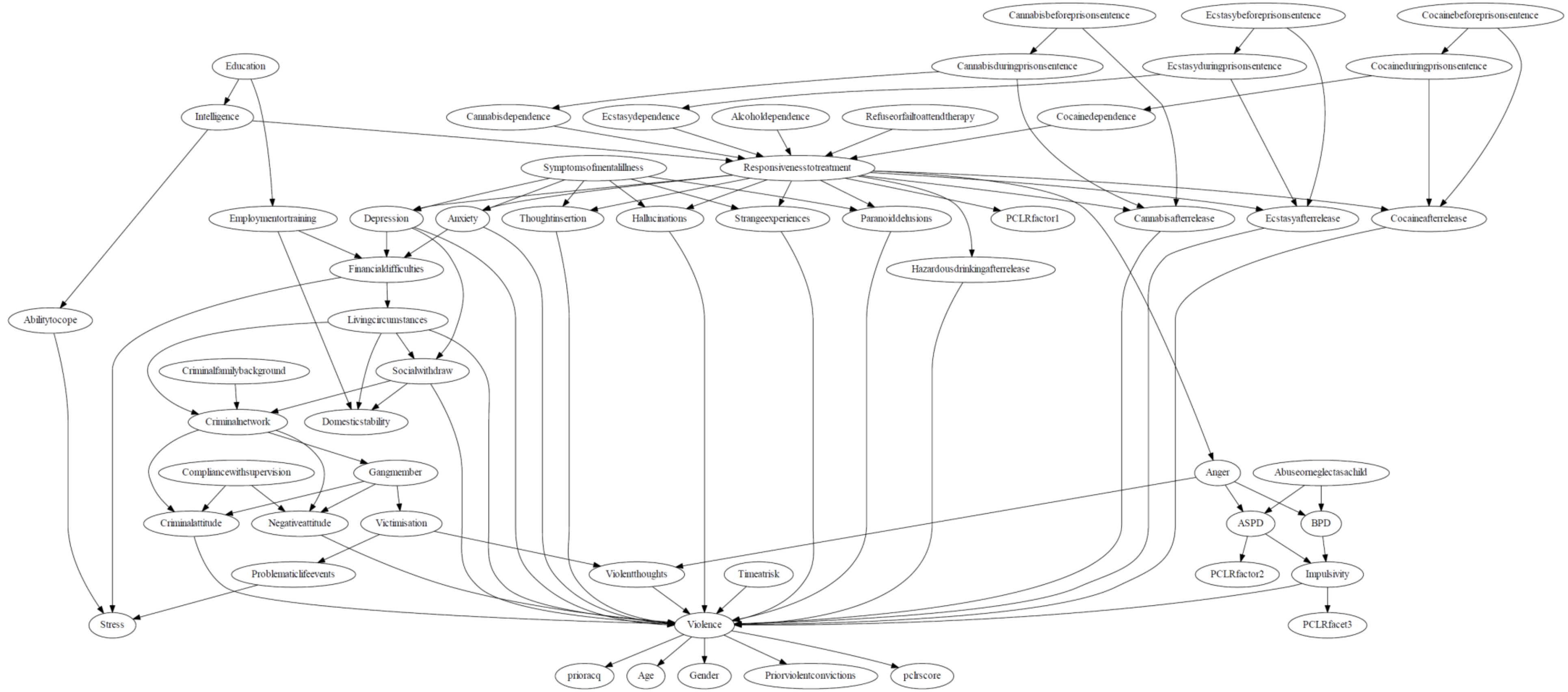

**Figure C.12.** The knowledge graph for case study *ForMed*.

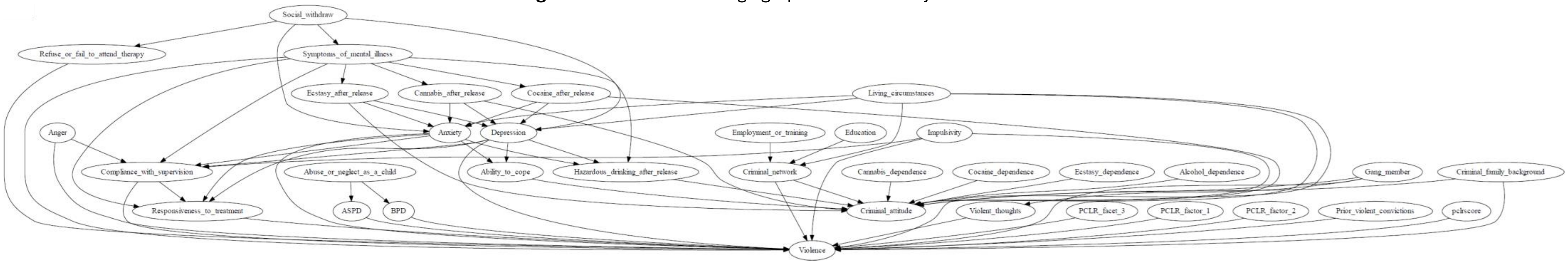

**Figure C.13.** The LLM (GPT-4) graph for case study *ForMed*.